\documentclass[runningheads]{llncs}
\usepackage{url}
\usepackage{multirow}

\usepackage{tikz}

\usepackage{bbding}

\usepackage{colortbl}

\usepackage{makecell}
\usepackage{bm}

\usepackage{amsmath}

\newcommand{\cubic}{Int.}
\newcommand{\mse}{$\mathcal{M}_{\rm B}$}
\newcommand{\raaa}{$\mathcal{M}_{+++}^R$}
\newcommand{\raab}{$\mathcal{M}_{++-}^R$}
\newcommand{\raba}{$\mathcal{M}_{+-+}^R$}
\newcommand{\rbaa}{$\mathcal{M}_{-++}^R$}
\newcommand{\rabb}{$\mathcal{M}_{+--}^R$}
\newcommand{\rbab}{$\mathcal{M}_{-+-}^R$}
\newcommand{\rbba}{$\mathcal{M}_{--+}^R$}
\newcommand{\rbbb}{$\mathcal{M}_{---}^R$}
\newcommand{\saaa}{$\mathcal{M}_{+++}^S$}
\newcommand{\saab}{$\mathcal{M}_{++-}^S$}
\newcommand{\saba}{$\mathcal{M}_{+-+}^S$}
\newcommand{\sbaa}{$\mathcal{M}_{-++}^S$}
\newcommand{\sabb}{$\mathcal{M}_{+--}^S$}
\newcommand{\sbab}{$\mathcal{M}_{-+-}^S$}
\newcommand{\sbba}{$\mathcal{M}_{--+}^S$}
\newcommand{\sbbb}{$\mathcal{M}_{---}^S$}

\begin{document}
\title{Task-driven real-world super-resolution of document scans}

\author{Maciej Zyrek\inst{1}, Tomasz Tarasiewicz\inst{1}, Jakub Sadel\inst{1}, Aleksandra Krzywon\inst{2}, and Michal Kawulok\inst{1}}
\institute{Department of Algorithmics and Software\\ Silesian University of Technology, Gliwice, Poland \\ \email{michal.kawulok@polsl.pl}
\and
Department of Biostatistics and Bioinformatics\\ Maria Sklodowska-Curie National Research Institute of Oncology, Gliwice Branch\\ Gliwice, Poland}



%
%
%
\maketitle              
\begin{abstract}
Single-image super-resolution refers to the reconstruction of a high-resolution image from a single low-resolution observation. Although recent deep learning-based methods have demonstrated notable success on simulated datasets---with low-resolution images obtained by degrading and downsampling high-resolution ones---they frequently fail to generalize to real-world settings, such as document scans, which are affected by complex degradations and semantic variability. In this study, we introduce a task-driven, multi-task learning framework for training a super-resolution network specifically optimized for optical character recognition tasks. We propose to incorporate auxiliary loss functions derived from high-level vision tasks, including text detection using the connectionist text proposal network, text recognition via a convolutional recurrent neural network, keypoints localization using Key.Net, and hue consistency. To balance these diverse objectives, we employ dynamic weight averaging mechanism, which adaptively adjusts the relative importance of each loss term based on its convergence behavior.  We validate our approach upon the SRResNet architecture, which is a well-established technique for single-image super-resolution. Experimental evaluations on both simulated and real-world scanned document datasets demonstrate that the proposed approach improves text detection, measured with intersection over union, while preserving overall image fidelity. These findings underscore the value of multi-objective optimization in super-resolution models for bridging the gap between simulated training regimes and practical deployment in real-world scenarios.

\end{abstract}

\section{Introduction}


Limited spatial resolution of scanned documents often poses a substantial challenge for optical character recognition (OCR) systems, in particular when input images suffer from the noise and other distortions~\cite{GuoDai2023} that result from sensor limitations, compression artifacts, motion blur, or suboptimal lighting conditions. 
In order to allow for effective processing of low-resolution (LR) scans with existing OCR systems, their quality can be enhanced with super-resolution (SR) techniques which operate either from a single image or multiple observations of the same scene. The state-of-the-art  single-image SR (SISR)  approaches are underpinned with deep learning. They include the first convolutional neural network (CNN) for SR (SRCNN)~\cite{Dong2014}, very deep SR network (VDSR)~\cite{Kim2016}, and a residual SR network SRResNet~\cite{Ledig2017} that was trained with a generative adversarial network (GAN) setting. These techniques, as well as recently-developed solutions~\cite{YuShi2024} demonstrate impressive performance on simulated benchmarks, with LR images obtained from original images that are later treated as HR references during training and validation. However, as noted by Cai et al.~\cite{CaiZeng2019RealSR}, these models often fail to generalize to real-world data where degradation is neither uniform nor well modeled by the kernels used for downsampling. This gave a rise to real-world SR that is aimed at super-resolving original rather than simulated images~\cite{ChenHe2022}. While such techniques consist in exploiting real-world datasets, they are rarely validated in task-specific scenarios~\cite{Kawulok2024JSTARS}.

To address these limitations, we propose a task-driven SISR framework designed specifically for document images. Our method builds on the SRResNet architecture, but shifts the training objective from pure image fidelity---for example, expressed with peak signal-to-noise ratio (PSNR)---to task-oriented performance. In particular, we introduce multiple auxiliary losses derived from pretrained semantic models. 
We incorporate four auxiliary loss components into our training framework: a text detection loss based on connectionist text proposal network (CTPN)~\cite{WangXie2019}, which encourages reconstruction of text-presence features; a loss derived from intermediate activations of a pretrained convolutional recurrent neural network (CRNN)~\cite{LiuWang2023,shi2016end}, promoting character recognition; a keypoint alignment loss using the Key.Net detector~\cite{Barroso-Laguna_2019_ICCV}, which enhances structural consistency between the output and ground truth; and a color consistency loss based on the hue component of the hue-saturation-value (HSV) color space that helps maintain chromatic coherence without distorting tonal relationships.

These complementary objectives are integrated into a unified multi-task loss function. A major challenge in such multi-objective optimization is determining appropriate weights for each task. Static weights may lead to unstable convergence or underfitting. To overcome this problem, we employ dynamic weight averaging (DWA)~\cite{LiuJohns2019}, which automatically adjusts task weights during training based on their relative learning dynamics. This encourages balanced training and ensures that no single loss component dominates the remaining ones.

\begin{figure}
    \centering
    \includegraphics[width=\textwidth]{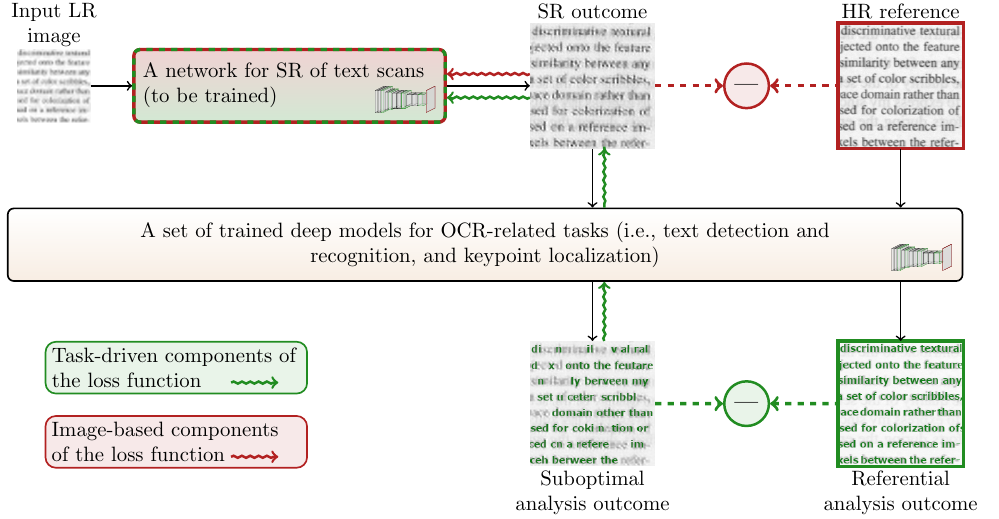}
    \caption{Outline of the proposed task-driven training of an SR network. CNNs performing OCR-related tasks are applied to process both the HR reference and the super-resolved image---the differences between their outcomes establish task-driven components of the loss function which are coupled with the commonly-applied image-driven components to guide the training of the SR network.}
    \label{fig:outline}
\end{figure}

To evaluate our approach, we have designed a comprehensive experimental setup involving real and simulated LR--HR document pairs. Real-world scans are acquired using controlled multi-resolution acquisition pipelines, while simulated pairs are generated through standard downsampling. We demonstrate that our task-driven training framework improves text detection performance measured with the intersection over union (IoU) metric, in particular for real-world scans. Importantly, the model shows strong generalization across different document types.
Overall, our contribution can summarized as follows:
\begin{enumerate}
 \item We propose a task-driven SISR framework (see Fig.~\ref{fig:outline}) aimed at super-resolving document images, guided by text detection and recognition, keypoint detection, and color consistency.
 \item We employ DWA to dynamically balance the loss components, ensuring stable and balanced convergence across the tasks.
 \item We introduce a real-world dataset with accurate registration at 4$\times$ magnification ratio, which allows for realistic evaluation of SR for OCR-related tasks.
 \item We provide a thorough analysis of model behavior on both real-world and simulated datasets, highlighting the practical benefits and risks of task-aware SR.
\end{enumerate}

The paper is structured as follows. In Section~\ref{sec:related}, we outline the state of the art in SR, with particular attention given to task-driven approaches. Our approach is specified in Section~\ref{sec:method} and the results of experimental validation are reported in Section~\ref{sec:exp}. Section~\ref{sec:concl} concludes the paper.

\section{Related Work} \label{sec:related}
In this section, we present the state of the art regarding SR techniques underpinned with deep learning (Section~\ref{sec:deepsr}), we outline the task-specific approaches, including super-resolving text documents (Section~\ref{sec:task-driven}), as well as the methods to handle multi-task optimization (Section~\ref{sec:multi-task}).


\subsection{Deep Learning for SR}\label{sec:deepsr}
Introduction of CNNs has revolutionized the field of SISR. The pioneering SRCNN model by Dong {et al.} optimized a pixel-wise $\rm L2$ loss to maximize PSNR on natural images~\cite{Dong2015}, and it was followed by the very deep VDSR network, which further improved reconstruction quality via residual learning~\cite{Kim2016}. Ledig {et al.\@} introduced SRResNet, a generator built from deep residual blocks, achieving state-of-the-art scores for Set5 and Set14 datasets~\cite{Ledig2017}. SRResNet also formed the backbone of SRGAN~\cite{Ledig2017}, which augments $\rm L2$ with an adversarial perceptual loss based on the features extracted using the VGG network~\cite{Simonyan2014}. Subsequent architectures such as enhanced deep SR network (EDSR)~\cite{LimSon2017} and a residual dense network (RDN)~\cite{Zhang2018} refined residual and attention mechanisms to push visual fidelity further. However, these methods focus primarily on natural-image benchmarks with synthetic downsampling and do not integrate higher-level semantic objectives during training. 

Recent developments in SR architectures increasingly emphasize reducing model size without compromising reconstruction quality~\cite{Ayazoglu2021}. Notably, recent work has shown that SISR can benefit from vision transformers~\cite{LuLi2022}, which dynamically adjust feature map sizes to lower model complexity. With just $6.5\cdot 10^5$ parameters, this approach has been reported to outperform many state-of-the-art models that are significantly larger. The swift parameter-free attention network~\cite{WanYu2024} introduces a novel attention mechanism devoid of parameters, striking a balance between image quality and inference speed, making it suitable for real-time applications. Xie {et al.} employed kernel distillation to simplify the model structure and enhance the attention modules~\cite{XieZhang2023}, reducing the computational cost while improving the performance.

\subsection{Task-Specific SR}\label{sec:task-driven}

Although most of the efforts are concerned with purpose-agnostic enhancement, validation of emerging SR techniques in the context of specific computer vision tasks receives growing research attention~\cite{Kawulok2024JSTARS}. This also includes research focused on text detection and recognition. Wang {et al.} used GAN-based SR to boost scene-text recognition performance~\cite{WangXie2019}, and Honda {et al.} leveraged multi-task transformers for scene-text SR~\cite{Honda2022Springer}. These methods typically report improvements in recognition accuracy. In the document domain, ICDAR Robust Reading Challenges have driven development of SR on scanned receipts and book pages~\cite{Gomez2017}. Inspired by these findings, our method explicitly incorporates OCR-relevant feature losses to guide SR toward text-clean reconstructions, while monitoring and mitigating potential hallucinations.

There were some attempts to train the networks for SR in a task-oriented manner by making the loss function focused on these specific tasks, thus guiding the training process accordingly. Haris {et al.} applied an object detection loss for training an SISR network~\cite{Haris2021}. Although the task-driven training leads to worse PSNR scores than relying on the $\rm L1$ loss, it was demonstrated that the super-resolved images constitute a more valuable source for object detection. Similar task-driven loss functions were also defined for semantic image segmentation~\cite{Frizza2022} and text recognition~\cite{Madi2022,WangXie2019}. Semantic segmentation was also used for learning an SR network in~\cite{Rad2020}. Importantly, all of these techniques were applied only to simulated LR images and they were not tested in real-world scenarios. This also concerns our recent work on task-driven SR~\cite{zyrek2024task} in which we trained several architectures for SISR in a task-oriented way. In the work reported here, we adapt that technique for real-world images and we incorporate multiple semantic losses that are dynamically balanced during training.


\subsection{Multi-Task Loss Optimization}\label{sec:multi-task}
Balancing heterogeneous loss terms is critical in multi-task learning. Fixed weights often cause one objective to dominate, leading to suboptimal convergence. Several dynamic strategies have been proposed in the literature~\cite{Vandenhende2021}, including {gradient normalization}~\cite{ChenBadrinarayanan2018} that equalizes gradient magnitudes across tasks, {uncertainty weighting}~\cite{Kendall2018} which uses task uncertainty as adaptive coefficients, dynamic task prioritization~\cite{GuoHaque2018}, and DWA~\cite{LiuJohns2019} which updates each weight based on recent loss change rates. DWA assigns greater emphasis to tasks whose losses decrease more slowly, promoting balanced convergence. 


\section{The Proposed Framework for Task-Driven Training} \label{sec:method}

Building upon the prior works reviewed in Section~\ref{sec:related}, in this section we present our method aimed at enhancing SISR for real-world scanned documents. While numerous studies have demonstrated promising performance in simulated settings, achieving robust and task-relevant reconstructions in real scenarios remains a significant challenge. Motivated by the observed limitations of conventional SISR techniques---particularly in preserving text-level semantics---we propose a task-driven, multi-loss training framework that we validate based on the SRResNet architecture. Our method integrates multiple objectives into the training pipeline to achieve reconstruction fidelity not only at the pixel level, but also in the semantic and structural domains critical to document analysis. 

The remainder of this section details our proposed solution. In Section~\ref{sec:network}, we describe the SRResNet architecture that we selected for our study along with the specific loss functions employed. Section~\ref{sec:loss} elaborates on our strategy for combining these objectives using DWA. 

\subsection{Network Architecture and Loss Function} \label{sec:network}

To address the challenge of super-resolving real-world document scans for OCR-related tasks, we propose a modular architecture centered around the SRResNet backbone~\cite{Ledig2017}, enriched with auxiliary components for semantic supervision. We have selected that network based on our earlier study~\cite{zyrek2024task} in which we considered several different architectures for SISR---SRResNet provides the optimal balance considering reconstruction accuracy and training time. Our design follows a task-driven paradigm in which reconstruction quality is not assessed solely via pixel-based similarity, but through its ability to retain task-relevant semantics essential for optical character recognition.

A key element of our framework is the integration of pretrained CTPN, CRNN, and Key.Net models into the loss function exploited for training the SR network. Importantly, the weights of these pretrained models remain frozen during training, so that they preserve their originally-trained behavior. The models for text detection (CTPN) and recognition (CRNN) were integrated into a single pipeline\footnote{The CTPN and CRNN models are available at \url{https://github.com/courao/ocr.pytorch}} that allows for solving the OCR task in an end-to-end manner~\cite{WangXie2019}. During training, these networks serve as \emph{proxy-supervisors}: rather than requiring manually annotated labels, we extract high-level feature activations from their deepest layers and compare them between the super-resolved image $\hat{I}$ and the ground-truth high-resolution image $I_{\mathrm{HR}}$ using the $\rm L1$ distance. In this way, each frozen network implicitly generates its own ``labels'' based on learned representations, yielding a semi-self-supervised training regime that enforces semantic consistency without explicit annotations.

Furthermore, we incorporate a pretrained Key.Net model (independent from OCR) to extract image keypoints, under the hypothesis that text regions exhibit distinctive keypoint patterns. Although Key.Net was originally designed for general-purpose keypoint detection, we test its ability to complement OCR-driven losses by encouraging structural alignment of salient features. These auxiliary loss functions guide the super-resolution model to preserve both textual and structural cues.

In the following subsections, we provide an overview of the core SRResNet model and describe the architectures and roles of the CTPN, CRNN, and Key.Net modules, which are integrated into the training pipeline to produce task-aligned super-resolution outputs. We focus specifically on SRResNet as the generator backbone, as this model demonstrated favorable results in our previous FedCSIS publication \cite{zyrek2024task}. Its balance between reconstruction quality, model complexity, and compatibility with multi-loss training made it a compelling choice for further task-driven adaptation.

\subsubsection{SRResNet Architecture Overview}

The SRResNet model, originally proposed by Ledig et al.~\cite{Ledig2017}, serves as the architectural backbone of our approach. The architecture is composed of an initial feature extraction layer that transforms the three-channel LR input image $I_{\mathrm{LR}} \in \mathcal{R}^{3 \times H \times W}$ into a higher-dimensional representation using a wide $9\times9$ convolution. This is followed by a deep stack of $N$ residual blocks, each of which refines the extracted features while maintaining stable training dynamics through skip connections. Formally, if $F_{i-1}$ is the input to the $i$-th residual block, then
\begin{equation}
F_i = F_{i-1} + \mathcal{R}(F_{i-1}), \quad i = 1,\dots,N,    
\end{equation}
where $\mathcal{R}(\cdot)$ denotes the two-layer convolutional transformation with parametric rectified linear unit (PReLU) activations. These residual connections allow the network to focus on learning the high-frequency difference between the HR and LR representations, rather than the entire mapping, thus improving convergence and final performance.

After the residual blocks, another convolutional layer merges the learned features into a single tensor, which is then upsampled by successive subpixel convolution (pixel shuffle) layers. Each subpixel block increases spatial resolution by a factor of 2$\times$ (for a total scaling factor of 4$\times$), rearranging feature-channel information into finer-grained pixel grids. Finally, a reconstruction layer (a wide $9\times9$ convolution followed by a Tanh activation) maps the upsampled feature maps back to a three-channel RGB image:
\begin{equation}
\hat{I} = G(I_{\mathrm{LR}}; \theta_G) \in \mathcal{R}^{3 \times (4H) \times (4W)}.
\end{equation}
In our work, $\theta_G$ denotes all trainable parameters of SRResNet. Because SRResNet has demonstrated strong baseline performance for text-centric SISR in prior work \cite{zyrek2024task}, we initialize it with the weights pretrained on the MS COCO dataset~\cite{WangXie2019} using a pure MSE loss before fine-tuning with our multi-task objectives.

\subsubsection{CTPN Architecture for Text Region Detection}

To extract spatial features relevant to textual content, we incorporate the CTPN network~\cite{WangXie2019}, originally introduced for detecting horizontal text lines in scene images. CTPN is particularly effective at identifying localized regions likely to contain text, making it a suitable auxiliary supervision module within our task-driven SR framework.

The architecture of the CTPN model is based on a truncated VGG16 backbone, pretrained on the ImageNet dataset~\cite{DengDong2009}. Given a super-resolved image $\hat{I}$ or an HR reference $I_{\mathrm{HR}}$, the convolutional encoder outputs a feature map of size $(C \times H' \times W')$. A subsequent lightweight $3 \times 3$ convolutional layer reduces the channel dimensionality to 512, resulting in a tensor $X \in \mathcal{R}^{512 \times H' \times W'}$. This tensor is reshaped into a sequence of vectors of length 512 and processed row-wise by a bidirectional gated recurrent unit (Bi-GRU), which captures horizontal contextual dependencies across the image width---essential for accurate text region detection. The output sequence, consisting of 256-dimensional features from the Bi-GRU, is reshaped back to a spatial map $X_{\mathrm{seq}} \in \mathcal{R}^{256 \times H' \times W'}$.

This intermediate representation is then passed through a $1 \times 1$ convolution to restore the dimensionality to 512 channels. The resulting tensor is processed by two parallel $1 \times 1$ convolutional branches. The first of these is a classification head, $\psi_{\mathrm{CTPN\text{-}clss}}$, which predicts, for each anchor (i.e., a fixed-width vertical slice of the input), the probability of containing text versus background. Training of this branch is carried out using masked cross-entropy loss, where neutral anchors are excluded from supervision. The second branch is the regression head, $\psi_{\mathrm{CTPN\text{-}reg}}$, which estimates the vertical coordinate offsets for the bounding box corresponding to each anchor. This output is trained with a Smooth $\rm L1$ loss applied only to positive samples. Through this design, the CTPN model captures both the presence and the precise location of textual content, enabling our SR framework to focus on semantically meaningful regions in the input images.

Concretely, if $F_{\mathrm{clss}}(\cdot)$ and $F_{\mathrm{reg}}(\cdot)$ denote the outputs of the classification and regression convolutions, then:
\begin{equation}
\mathcal{L}_{\mathrm{CTPN\text{-}clss}} = \mathrm{CrossEntropy}\bigl(F_{\mathrm{clss}}(\hat{I}),\,F_{\mathrm{clss}}(I_{\mathrm{HR}})\bigr) \rm ,
\end{equation}
\begin{equation}
\mathcal{L}_{\mathrm{CTPN\text{-}reg}} = \mathrm{SmoothL1}\bigl(F_{\mathrm{reg}}(\hat{I}),\,F_{\mathrm{reg}}(I_{\mathrm{HR}})\bigr) \rm .
\end{equation}
When used as a proxy for network supervision, however, we do not directly compare the final predicted boxes. Instead, we extract deep features from the last pre-activation layers in each head—denoted $\psi_{\mathrm{CTPN\text{-}deep}}(\cdot)\in\mathcal{R}^D$—and measure their $\rm L1$ distance between $\hat{I}$ and $I_{\mathrm{HR}}$:
\begin{equation}
\mathcal{L}_{\mathrm{CTPN\text{-}deep}} 
= \bigl\lVert \psi_{\mathrm{CTPN\text{-}deep}}(\hat{I}) - \psi_{\mathrm{CTPN\text{-}deep}}(I_{\mathrm{HR}}) \bigr\rVert_1.
\end{equation}
These frozen-features losses compel the SR network to preserve text-proposal semantics in critical regions.

\subsubsection{CRNN Architecture for Text Recognition}

To provide high-level semantic supervision aligned with text recognition, we integrate a frozen CRNN module~\cite{shi2016end} into our training framework, trained on a large corpus of scanned documents. It remains frozen throughout SR training and serves as a proxy for text-recognition features.

CRNN consists of a convolutional encoder that transforms an RGB input into a compact feature map whose height is reduced to 1, effectively converting the 2D image into a 1D sequence of feature vectors along the width axis. This convolutional encoder comprises several stacked convolutional layers with BatchNorm and rectified linear unit (ReLU) and pooling layers that progressively halve the height dimension until it equals one. The output of this stage is $H \in \mathcal{R}^{C \times 1 \times W'}$, which is reshaped into a sequence $S \in \mathcal{R}^{W' \times C}$.

Next, $S$ is fed into two stacked bidirectional long short-term memory (Bi-LSTM) layers. Each Bi-LSTM layer processes the sequence in both forward and backward directions, capturing context across neighboring character regions. The final recurrent output is projected via a linear embedding to a distribution over character classes plus a blank token. In our multi-loss setup, we  extract intermediate recurrent features denoted as $\psi_{\mathrm{CRNN}}(\cdot)\in\mathcal{R}^{W'\times d}$ from the last Bi-LSTM layer and compare them between $\hat{I}$ and $I_{\mathrm{HR}}$:
\begin{equation}
\mathcal{L}_{\mathrm{CRNN}} 
= \bigl\lVert \psi_{\mathrm{CRNN}}(\hat{I}) - \psi_{\mathrm{CRNN}}(I_{\mathrm{HR}}) \bigr\rVert_1.
\end{equation}
This encourages the SR model to produce features that match those of the HR reference, thus preserving character-level discriminability.

\subsubsection{Key.Net for Structural Consistency}

While CTPN and CRNN focus on OCR-driven supervision, we also incorporate Key.Net~\cite{Barroso-Laguna_2019_ICCV}, a pretrained network for generic keypoint detection in natural images. Our hypothesis is that text regions induce characteristic local keypoints (e.g., stroke junctions), so aligning these keypoints between $\hat{I}$ and $I_{\mathrm{HR}}$ may further reinforce structural fidelity. Key.Net is sourced from its original implementation and remains frozen.

Given an image $X$, let $\mathbf{p}_k(X)\in\mathcal{R}^2$ denote the 2D coordinates (or heatmap response) of the $k$-th detected keypoint. We then define the keypoint alignment loss as:
\begin{equation}
\mathcal{L}_{\mathrm{Key.Net}} 
= \sum_{k} \bigl\lVert \mathbf{p}_k(\hat{I}) - \mathbf{p}_k(I_{\mathrm{HR}}) \bigr\rVert_2^2,
\end{equation}
summing over all keypoints detected in the HR image. In practice, we sample a fixed number of top-scoring keypoints from Key.Net and enforce that their spatial arrangement be preserved after SR.

\subsubsection{Multi‐Losses Approach}


In the study reported here, we consider eight loss components to capture pixel-level fidelity, structural consistency, and high-level semantic alignment. These are: (\textit{i})~{pixel-wise MSE}, (\textit{ii})~consistency loss component, (\textit{iii})~distances in the CTPN feature spaces (three components), (\textit{iv})~distance in the CRNN feature space, (\textit{v})~the Key.Net loss, and (vi)~the hue difference. 

The pixel-wise MSE loss enforces that the super-resolved output matches the HR ground truth in an $\rm L2$ sense. Formally:
\begin{equation}
\mathcal{L}_{\mathrm{MSE}} \;=\; \bigl\lVert \hat{I} - I_{\mathrm{HR}} \bigr\rVert_2^2.
\end{equation}
Minimizing $\mathcal{L}_{\mathrm{MSE}}$ drives the network to reduce per‐pixel differences, which typically correlates with high PSNR values.

The goal of the consistency loss is to ensure that when a super-resolved image is downsamped back to the original size, it resembles the input image. By bicubically downsampling $\hat{I}$ back to the LR domain and comparing it to the original $I_{\mathrm{LR}}$, we impose the cycle-consistency:
\begin{equation}
\mathcal{L}_{\mathrm{cons}} \;=\; \bigl\lVert D(\hat{I}) - I_{\mathrm{LR}} \bigr\rVert_2^2,
\end{equation}
where $D(\cdot)$ denotes bicubic downsampling by a factor of 4$\times$. This term penalizes unrealistic high-frequency artifacts that cannot be justified by the original LR image.


To guide the network toward preserving spatial features relevant to text detection, we incorporate a set of auxiliary loss functions derived from a frozen CTPN model. Specifically, we extract and compare three types of intermediate representations from both the super-resolved image $\hat{I}$ and its HR counterpart $I_{\mathrm{HR}}$: (\textit{i})~deep convolutional features, (\textit{ii})~classification logits, and (\textit{iii})~regression outputs. Each feature space yields a separate $\rm L1$ loss, formulated as:
\begin{equation}
\mathcal{L}_{\mathrm{CTPN\text{-}deep}} = \bigl\lVert\,\psi_{\mathrm{CTPN\text{-}deep}}(\hat{I}) - \psi_{\mathrm{CTPN\text{-}deep}}(I_{\mathrm{HR}})\bigr\rVert_1,
\end{equation}
\begin{equation}
\mathcal{L}_{\mathrm{CTPN\text{-}clss}} = \bigl\lVert\,\psi_{\mathrm{CTPN\text{-}clss}}(\hat{I}) - \psi_{\mathrm{CTPN\text{-}clss}}(I_{\mathrm{HR}})\bigr\rVert_1,
\end{equation}
\begin{equation}
\mathcal{L}_{\mathrm{CTPN\text{-}reg}}  = \bigl\lVert\,\psi_{\mathrm{CTPN\text{-}reg}}(\hat{I})  - \psi_{\mathrm{CTPN\text{-}reg}}(I_{\mathrm{HR}})\bigr\rVert_1.
\end{equation}
Here, $\psi_{\mathrm{CTPN\text{-}deep}}(\cdot)$ denotes the deepest convolutional activation maps, $\psi_{\mathrm{CTPN\text{-}clss}}(\cdot)$ refers to the classification output logits for text proposal confidence, and $\psi_{\mathrm{CTPN\text{-}reg}}(\cdot)$ captures the predicted vertical bounding-box regressions.

By encouraging alignment between these intermediate representations extracted from $\hat{I}$ and $I_{\mathrm{HR}}$, we incentivize the network to reconstruct text regions in a manner that retains their original spatial structure and saliency. This formulation provides indirect supervision for text-relevant geometry, even in the absence of explicit bounding-box labels.

To further ensure that the super-resolved output retains character-level recognizability, we incorporate an auxiliary loss based on feature activations from the pretrained CRNN model~\cite{shi2016end}. Let $\psi_{\mathrm{CRNN}}(\cdot)$ denote the feature extraction function of the CRNN model, which includes both convolutional and recurrent representations. We compute the $\rm L1$ distance between feature maps extracted from the super-resolved image $\hat{I}$ and its HR reference $I_{\mathrm{HR}}$:
\begin{equation}
\mathcal{L}_{\mathrm{CRNN}} \;=\; \bigl\lVert\,\psi_{\mathrm{CRNN}}(\hat{I}) - \psi_{\mathrm{CRNN}}(I_{\mathrm{HR}})\bigr\rVert_1.
\end{equation}
This formulation encourages the network to reconstruct images that preserve not only low-level textures but also the semantic structure required for accurate text transcription. By aligning CRNN-derived features across the input pairs, the model is implicitly guided to maintain OCR-relevant details even in the absence of explicit textual labels.


To encourage the preservation of structural patterns and spatial consistency in super-resolved outputs, we incorporate the {multi-scale index proposal (MSIP) loss derived from the Key.Net architecture. Unlike previous losses that compare intermediate feature maps, MSIP operates directly on local descriptors extracted from keypoint regions across multiple scales. 
Given the HR reference image $I_{\mathrm{HR}}$ and the super-resolved image $\hat{I}$, Key.Net identifies repeatable keypoints and computes local gradient-based descriptors. The MSIP loss compares the responses around corresponding keypoints at multiple scales, enforcing stability and robustness in the reconstructed geometry. Formally, this loss is defined as:
\begin{equation}
\mathcal{L}_{\mathrm{Key.Net}} \;=\; \sum_{s \in \mathcal{S}} \sum_{k} \bigl\lVert\, \phi_{s,k}(\hat{I}) - \phi_{s,k}(I_{\mathrm{HR}}) \bigr\rVert_2^2,
\end{equation}
where $\phi_{s,k}(\cdot)$ denotes the local descriptor at keypoint $k$ and scale $s$, and $\mathcal{S}$ is the set of scales used. This loss penalizes discrepancies in the geometric structure and ensures that the super-resolved image maintains stable keypoint representations across multiple resolutions.

By leveraging the MSIP loss, we promote alignment not only in texture or color, but in intrinsic structural cues, which are critical in documents containing fine typographic details or graphical annotations.


The overall objective function is defined as a weighted combination of all individual loss components described previously. This aggregation allows the network to simultaneously optimize for low-level fidelity, structural alignment, semantic consistency, and chromatic coherence. To ensure that no single task dominates the training process, each weight $\lambda_i$ is dynamically adjusted during optimization using the DWA strategy~\cite{Liu2019}. Specifically, higher emphasis is placed on tasks whose loss decreases more slowly over time, thus promoting balanced convergence across heterogeneous objectives. The total loss is computed as follows:
\begin{align}
\label{eq:total_loss}
\mathcal{L}_{\mathrm{total}} =\;& 
\lambda_{\mathrm{MSE}}\,\mathcal{L}_{\mathrm{MSE}} \;+\;
\lambda_{\mathrm{cons}}\,\mathcal{L}_{\mathrm{cons}} \;+\;
\lambda_{\mathrm{CTPN\text{-}deep}}\,\mathcal{L}_{\mathrm{CTPN\text{-}deep}} \nonumber \\
&\quad +\;\lambda_{\mathrm{CTPN\text{-}clss}}\,\mathcal{L}_{\mathrm{CTPN\text{-}clss}} \;+\;
\lambda_{\mathrm{CTPN\text{-}reg}}\,\mathcal{L}_{\mathrm{CTPN\text{-}reg}} \nonumber \\
&\quad +\;\lambda_{\mathrm{CRNN}}\,\mathcal{L}_{\mathrm{CRNN}} \;+\;
\lambda_{\mathrm{Key.Net}}\,\mathcal{L}_{\mathrm{Key.Net}} \;+\;
\lambda_{\mathrm{Hue}}\,\mathcal{L}_{\mathrm{Hue}}.
\end{align}

This formulation ensures that the super-resolved output not only minimizes pixel-wise reconstruction error, but also retains structural and semantic integrity relevant for downstream document analysis tasks such as text detection and recognition. The DWA mechanism is recomputed at the end of each epoch, based on the relative loss descent rates, thereby allowing the model to adapt its focus throughout the training trajectory. 
The weights $\{\lambda_i(t)\}$ are updated at each epoch $t$. Let 
\begin{equation}
r_i(t) = \frac{\mathcal{L}_i(t-1)}{\mathcal{L}_i(t-2)}
\end{equation}
denote the relative loss improvement between two earlier epochs. Then:
\begin{equation}
\lambda_i(t) = \frac{N \exp\bigl(r_i(t-1) / T\bigr)}{\sum_{j=1}^N \exp\bigl(r_j(t-1) / T\bigr)},
\end{equation}
where $N=8$ is the number of loss terms and $T=2$ controls the weight softness. This encourages balanced convergence, preventing any single task from dominating training.

\subsection{Multi-Task Loss Aggregation and Optimization Strategy}\label{sec:loss}

The loss functions described above are combined into a unified multi-task training objective, where each component contributes to the overall optimization according to its dynamically-updated weight. This composite strategy enables the model to simultaneously satisfy multiple reconstruction goals—ranging from pixel-wise accuracy to high-level semantic preservation—during training.


All loss values are aggregated into the total loss:
\begin{equation}
\label{eq:final_total_loss}
\mathcal{L}_{\mathrm{total}} = \sum_{i=1}^{N} \lambda_i(t) \cdot \mathcal{L}_i,
\end{equation}
and gradients are propagated accordingly to update the SRResNet parameters $\theta$ using the Adam optimizer.

\section{Experimental Validation} \label{sec:exp}

In this section, we report our experiments conducted to evaluate the proposed task-driven SR framework. We describe the dataset construction and training setup (Section~\ref{sec:setup}), followed by quantitative and qualitative comparisons across multiple test scenarios (Section~\ref{sec:results}), and the final discussion (Section~\ref{sec:discussion}). 

\subsection{Experimental Setup} \label{sec:setup}

To evaluate the effectiveness of our approach, we carried out a series of controlled experiments comparing models trained under various supervision regimes. Each model was evaluated on a diverse set of test datasets, with both genuine and simulated degradations, to assess its generalization ability and robustness to real-world distortions. In what follows, we describe the dataset preparation procedure, the training configuration, and the evaluation protocol used in our study.

\subsubsection{Datasets}

To evaluate the effectiveness of our task-driven SISR framework, we constructed a high-quality dataset of real document scans acquired under different conditions\footnote{We will publish the dataset upon paper acceptance}. The dataset design was guided by prior studies and continues the line of investigation from our earlier work, focusing on the magnification factor of $\times4$. This upsampling factor offers a practical balance: while $2\times$ magnification may not sufficiently challenge the model to reconstruct fine details, higher ratios often lead to unstable optimization and hallucinated content. Thus, the $4\times$ setting provides a meaningful and realistic benchmark for restoration quality. The dataset is composed of several parts, namely: (\textit{i})~University Bulletin (scans of a university bulletin), (\textit{ii})~Scientific Article (scans of a printed scientific article), (\textit{iii})~COVID Test Leaflet (a scan of a medical leaflet), as well as (\textit{iv})~a publically available Old Books dataset~\cite{Barcha2018}. Furthermore, we have exploited the MS COCO dataset~\cite{WangXie2019} that contains natural images.  

The university bulletin and scientific article scans were acquired using a Samsung SCX-3400 flatbed scanner---a widely available consumer-grade device. Each page was scanned sequentially at five resolution levels: 75, 150, 200, 300, and 600 dots per inch (DPI). The scanning was automated via a custom script to maintain consistent acquisition conditions. After scanning at all DPI settings, each page was translated by a small amount and re-scanned, yielding nine spatial shifts per page. This resulted in a total of $32 \times 5 \times 9 = 1440$ scanned images. The resulting multi-resolution, multi-shift dataset supports both single-image and multi-frame SR scenarios. For our study, we focus on resolution pairs with a scaling ratio of exactly $4\times$ (75--300\,DPI and 150--600\,DPI), ensuring both practical fidelity and alignment with our experimental goals.

To ensure spatial alignment between LR and HR image pairs, we applied global registration based on rigid 2D translation. Each LR image was first upsampled using bicubic interpolation to match the resolution of its HR counterpart. To assess alignment quality, we computed an absolute difference map, followed by calculation of the MSE within a central crop region (excluding a 32-pixel margin). A stochastic grid search was then performed to identify the integer-pixel translation vector that minimized the MSE. To ensure the reliability of the estimated transformation, we validated the chosen shift across 20 randomly selected image pairs. The final translations were determined to be $[-5, -1]$ for 75--300\,DPI pairs and $[-6, -7]$ for 150--600\,DPI pairs. These displacements were subsequently applied to the LR images using affine transformation. Aligned image pairs were cropped into patch pairs of $256\times256$ pixels (LR) and $1024\times1024$ pixels (HR), preserving the $4\times$ ratio. The stride matched the LR patch size, and patches were clamped at image borders to avoid out-of-bound sampling. For each page, the zero-shift scan was used as the validation example for all DPI levels. The remaining eight shifted scans served as the training set, yielding an 8:1 split per resolution. 

The COVID test leaflet was scanned using a consumer-grade Canon imageFORMULA P-208II linear scanner equipped with a contact image sensor. These scans pose significant challenges due to transport instabilities during scanning, which result in nonlinear geometric distortions such as local stretching and warping. Consequently, this dataset provides a robust benchmark for assessing the spatial resilience of SR models under realistic and uncontrolled acquisition conditions. Each scan was divided into overlapping tiles with $512 \times 512$ pixels. Local alignment was performed using phase correlation, and the registered tiles were reassembled into full images. Due to the non-rigid nature of distortions, all registration results were manually verified. These scans were incorporated into the test set.

The {Old Books} dataset~\cite{Barcha2018} is publically available and comprises HR scans of historical printed documents. It features a wide variety of typographic styles, physical degradation due to paper aging, and uneven illumination. Importantly, the dataset includes text transcriptions and binarized ground truth versions of each scan, making it particularly suitable for evaluating OCR-related performance on super-resolved outputs.

To examine generalization beyond document-centric imagery, we employ a simulated subset of the MS COCO dataset~\cite{WangXie2019}. From this corpus, we extracted image regions that contain textual content and generated corresponding LR images via bicubic downsampling. This controlled setup enables benchmarking of our document-oriented SR model against natural-scene inputs, offering insights into its cross-domain adaptability.









For the COVID Bulletin and Old Books datasets, we prepare both real LR inputs obtained from physical scans and synthetic LR--HR pairs generated by downsampling the corresponding HR images. This dual-mode evaluation facilitates fair and consistent comparison across different degradation regimes, helping to disentangle the effects of acquisition noise from purely resolution-related loss. In contrast, the MS COCO subset is included solely into the simulated dataset, as no real LR counterparts are available. 

\subsubsection{Training Strategy}

The SRResNet model was initialized using weights from a variant pretrained on the MS COCO dataset with pixel-wise MSE loss, following the procedure described in~\cite{Ledig2017}. To investigate the effect of degradation realism on SR performance, we adopted a two-stage fine-tuning strategy that separates the impact of real and simulated data conditions.

In the first stage, referred to as real-data fine-tuning, we trained the model using image pairs acquired through actual scanning processes, specifically at resolution ratios of $4\times$ (e.g., 75--300 DPI and 150--600 DPI). These pairs inherently contain realistic distortions, including scanner-specific blur, noise, compression artifacts, and physical imperfections from printed media. This setup enables the model to learn reconstruction patterns reflective of real-world scanning degradations.

In the second stage, termed synthetic-data fine-tuning, we reused the same HR images and applied bicubic downsampling to generate their LR counterparts. This allowed for controlled degradation modeling under matched semantic content, isolating the learning dynamics attributable solely to the downsampling process. This two-pass protocol was designed to disentangle the contributions of physical acquisition artifacts and interpolation-based degradation, facilitating a fair comparison between models trained under real and synthetic conditions.

To balance the multiple objectives of our task-driven framework, we employed the DWA strategy for adaptive loss weighting. 
In practice, we observed that DWA allowed the model to dynamically shift focus depending on input characteristics: for instance, when processing patches with sparse or low-contrast text, the contribution of CRNN-based supervision increased, whereas for texture-rich regions, greater emphasis was placed on pixel-level losses and Key.Net-based structural alignment. This adaptive mechanism promoted stable optimization and contributed to the model’s ability to generalize effectively across the test dataset.

\subsubsection{Investigated Variants and Evaluation Metrics} 

In Table~\ref{tab:variants}, we enlist the variants investigated within our study. As it was demonstrated (for simulated datasets) that CTPN losses are crucial for optimizing SR networks for OCR-related tasks, here we focus on investigating the influence of the components concerned with CRNN, Key.Net, and Hue features. In addition to that, we report the results obtained with bicubic interpolation (\cubic) and with a baseline SRResNet model trained from the simulated MS COCO dataset (\mse). 
\begin{table}
    \centering
        \caption{The training setups considered in our experimental study.}

    \begin{tabular}{rlccc}
    
    \Xhline{2\arrayrulewidth}
    Variant name & Training dataset & $\mathcal{L}_{\mathrm{CRNN}}$ & $\mathcal{L}_{\mathrm{Key.Net}}$ & $\mathcal{L}_{\mathrm{Hue}}$ \\ \hline
\raaa & Real-world & \Checkmark & \Checkmark & \Checkmark \\
\raab & Real-world & \Checkmark & \Checkmark &  \\
\raba & Real-world & \Checkmark &  & \Checkmark \\
\rabb & Real-world & \Checkmark &  &  \\
\rbaa & Real-world &  & \Checkmark & \Checkmark \\
\rbab & Real-world &  & \Checkmark &  \\
\rbba & Real-world &  &  & \Checkmark \\
\rbbb & Real-world &  &  &  \\
\saaa & Simulated & \Checkmark & \Checkmark & \Checkmark \\
\saab & Simulated & \Checkmark & \Checkmark &  \\
\saba & Simulated & \Checkmark &  & \Checkmark \\
\sabb & Simulated & \Checkmark &  &  \\
\sbaa & Simulated &  & \Checkmark & \Checkmark \\
\sbab & Simulated &  & \Checkmark &  \\
\sbba & Simulated &  &  & \Checkmark \\
\sbbb & Simulated &  &  &  \\
    \Xhline{2\arrayrulewidth}
    \end{tabular}
    \label{tab:variants}
\end{table}

For each variant, we report the PSNR, structural similarity index (SSIM), learned perceptual image patch similarity (LPIPS), and intersection over union (IoU) between the text localization extracted from super-resolved and HR reference images. In order to verify whether the differences between the investigated variants are statistically significant, we have employed statistical tests. Comparisons between groups were made using the Kruskal-Wallis H test and the post hoc Dunn test with Benjamini-Hochberg $p$-value correction. A two‐sided $p$-value < 0.05 was considered statistically significant. All computational analysis was performed in the R environment for statistical computing (version 4.4.3).

\subsection{Results} \label{sec:results}

The reconstruction quality scores obtained using the investigated variants over the test datasets are reported in Table~\ref{tab:scores}. It can be seen that all of the trained SR models retrieve worse image fidelity scores (PSNR, SSIM and LPIPS) than bicubic interpolation, but the IoU scores are better relying on the models trained in a task-driven way from real-world datasets (both for simulated and real-world datasets). The violin plots showing distribution of the scores for the simulated and real world datasets are presented in Figures~\ref{fig:plot_iou_sim} and~\ref{fig:plot_iou_real}, respectively. In Tables~\ref{tab:psnr_sim}--\ref{tab:iou_real}, we report the outcomes of statistical tests. It can be seen that the IoU scores are significantly better for the models trained in a task-driven way both for the simulated (Table~\ref{tab:iou_sim}) and real-world (Table~\ref{tab:iou_real}) datasets.
\begin{table}[t]
    \centering

\caption{Reconstruction quality scores (mean value and standard deviation) retrieved using different models for simulated and real-world datasets. For each metric, we show whether higher  ($\uparrow$) or lower  ($\downarrow$) scores are better. The best scores for each metric are boldfaced.}

\resizebox{\textwidth}{!}{
    \begin{tabular}{lccccccccc}

    \Xhline{2\arrayrulewidth}
    
    & \multicolumn{4}{c}{Simulated images} & & \multicolumn{4}{c}{Real-world images} \\ \cline{2-5} \cline{7-10}
        & PSNR [dB] $\uparrow$ & SSIM $\uparrow$ & LPIPS $\downarrow$ & IoU $\uparrow$ & & PSNR [dB] $\uparrow$ & SSIM $\uparrow$ & LPIPS $\downarrow$ & IoU $\uparrow$ \\ \hline

\cubic & 	$ \bm{20.36\pm 6.03}$ & 	$ \bm{0.7043\pm 0.1890}$ & 	$ \bm{0.2973\pm 0.1893}$ & 	$ 0.9038\pm 0.0993$ & 	&	$ \bm{19.02\pm 6.32}$ & 	$ \bm{0.7403\pm 0.1934}$ & 	$ \bm{0.2516\pm 0.1863}$ & 	$ 0.8820\pm 0.0961$ \\[0.5mm]
\mse & 	$ 18.09\pm 5.35$ & 	$ 0.6535\pm 0.1926$ & 	$ 0.3222\pm 0.1914$ & 	$ 0.9033\pm 0.0943$ & 	&	$ 15.67\pm 5.39$ & 	$ 0.6509\pm 0.2182$ & 	$ 0.3073\pm 0.2074$ & 	$ 0.8703\pm 0.0960$ \\[0.5mm]
\raaa & 	$ 15.52\pm 5.31$ & 	$ 0.4512\pm 0.2423$ & 	$ 0.4186\pm 0.1976$ & 	$ 0.9119\pm 0.1029$ & 	&	$ 13.72\pm 5.20$ & 	$ 0.4182\pm 0.2698$ & 	$ 0.4424\pm 0.2139$ & 	$ \bm{0.8894\pm 0.0979}$ \\[0.5mm]
\raab & 	$ 17.62\pm 3.98$ & 	$ 0.6195\pm 0.1995$ & 	$ 0.3083\pm 0.1541$ & 	$ 0.9071\pm 0.1022$ & 	&	$ 17.19\pm 4.38$ & 	$ 0.6708\pm 0.1892$ & 	$ 0.2799\pm 0.1471$ & 	$ 0.8834\pm 0.1005$ \\[0.5mm]
\raba & 	$ 17.39\pm 3.86$ & 	$ 0.6290\pm 0.2014$ & 	$ 0.3032\pm 0.1608$ & 	$ 0.9094\pm 0.0989$ & 	&	$ 16.74\pm 4.04$ & 	$ 0.6793\pm 0.1859$ & 	$ 0.2723\pm 0.1544$ & 	$ 0.8823\pm 0.1027$ \\[0.5mm]
\rabb & 	$ 14.93\pm 4.80$ & 	$ 0.4990\pm 0.2206$ & 	$ 0.4071\pm 0.1894$ & 	$ 0.8904\pm 0.1191$ & 	&	$ 13.97\pm 5.14$ & 	$ 0.4901\pm 0.2579$ & 	$ 0.4116\pm 0.2206$ & 	$ 0.8635\pm 0.1233$ \\[0.5mm]
\rbaa & 	$ 16.12\pm 4.55$ & 	$ 0.5360\pm 0.2069$ & 	$ 0.3471\pm 0.1637$ & 	$ \bm{0.9142\pm 0.1041}$ & 	&	$ 14.65\pm 4.70$ & 	$ 0.5214\pm 0.2435$ & 	$ 0.3619\pm 0.1784$ & 	$ 0.8876\pm 0.0990$ \\[0.5mm]
\rbab & 	$ 16.77\pm 4.10$ & 	$ 0.4628\pm 0.1842$ & 	$ 0.3760\pm 0.1488$ & 	$ 0.9038\pm 0.1112$ & 	&	$ 15.42\pm 4.38$ & 	$ 0.4364\pm 0.2158$ & 	$ 0.3923\pm 0.1711$ & 	$ 0.8776\pm 0.1041$ \\[0.5mm]
\rbba & 	$ 17.44\pm 4.17$ & 	$ 0.6053\pm 0.1793$ & 	$ 0.3204\pm 0.1549$ & 	$ 0.9123\pm 0.0959$ & 	&	$ 16.24\pm 4.21$ & 	$ 0.6417\pm 0.1728$ & 	$ 0.2965\pm 0.1555$ & 	$ 0.8873\pm 0.0968$ \\[0.5mm]
\rbbb & 	$ 14.90\pm 4.87$ & 	$ 0.5027\pm 0.2197$ & 	$ 0.4085\pm 0.1785$ & 	$ 0.8795\pm 0.1335$ & 	&	$ 13.99\pm 5.30$ & 	$ 0.5016\pm 0.2536$ & 	$ 0.4052\pm 0.2050$ & 	$ 0.8549\pm 0.1386$ \\[0.5mm]
\saaa & 	$ 15.95\pm 5.40$ & 	$ 0.5516\pm 0.2094$ & 	$ 0.3239\pm 0.1896$ & 	$ 0.9062\pm 0.0883$ & 	&	$ 13.58\pm 5.45$ & 	$ 0.5229\pm 0.2530$ & 	$ 0.3473\pm 0.2114$ & 	$ 0.8692\pm 0.1036$ \\[0.5mm]
\saab & 	$ 16.57\pm 5.03$ & 	$ 0.5468\pm 0.2057$ & 	$ 0.3260\pm 0.1803$ & 	$ 0.9010\pm 0.0892$ & 	&	$ 14.26\pm 4.75$ & 	$ 0.5168\pm 0.2457$ & 	$ 0.3591\pm 0.2038$ & 	$ 0.8662\pm 0.1033$ \\[0.5mm]
\saba & 	$ 16.42\pm 4.53$ & 	$ 0.5808\pm 0.1796$ & 	$ 0.3253\pm 0.1642$ & 	$ 0.9028\pm 0.0903$ & 	&	$ 14.38\pm 4.27$ & 	$ 0.5715\pm 0.2046$ & 	$ 0.3406\pm 0.1780$ & 	$ 0.8682\pm 0.1023$ \\[0.5mm]
\sabb & 	$ 16.67\pm 5.19$ & 	$ 0.5463\pm 0.2181$ & 	$ 0.3217\pm 0.1785$ & 	$ 0.9006\pm 0.0894$ & 	&	$ 14.41\pm 4.98$ & 	$ 0.5193\pm 0.2541$ & 	$ 0.3466\pm 0.2007$ & 	$ 0.8658\pm 0.1015$ \\[0.5mm]
\sbaa & 	$ 16.64\pm 4.59$ & 	$ 0.5925\pm 0.1691$ & 	$ 0.3224\pm 0.1595$ & 	$ 0.9020\pm 0.0876$ & 	&	$ 14.56\pm 4.30$ & 	$ 0.5848\pm 0.1901$ & 	$ 0.3444\pm 0.1734$ & 	$ 0.8663\pm 0.1030$ \\[0.5mm]
\sbab & 	$ 16.58\pm 5.07$ & 	$ 0.5505\pm 0.2075$ & 	$ 0.3253\pm 0.1787$ & 	$ 0.9007\pm 0.0881$ & 	&	$ 14.37\pm 4.83$ & 	$ 0.5250\pm 0.2459$ & 	$ 0.3564\pm 0.2005$ & 	$ 0.8652\pm 0.1020$ \\[0.5mm]
\sbba & 	$ 17.26\pm 4.79$ & 	$ 0.6069\pm 0.1737$ & 	$ 0.3058\pm 0.1577$ & 	$ 0.9015\pm 0.0933$ & 	&	$ 15.58\pm 4.77$ & 	$ 0.6049\pm 0.1949$ & 	$ 0.3104\pm 0.1655$ & 	$ 0.8664\pm 0.1045$ \\[0.5mm]
\sbbb & 	$ 16.71\pm 5.17$ & 	$ 0.5609\pm 0.2096$ & 	$ 0.3163\pm 0.1730$ & 	$ 0.9007\pm 0.0874$ & 	&	$ 14.49\pm 5.00$ & 	$ 0.5429\pm 0.2417$ & 	$ 0.3357\pm 0.1909$ & 	$ 0.8652\pm 0.1006$ \\[0.5mm]

\Xhline{2\arrayrulewidth}
         
    \end{tabular}
    }
    
    \label{tab:scores}
\end{table}

\begin{figure}
    \centering
    \includegraphics[width=\textwidth]{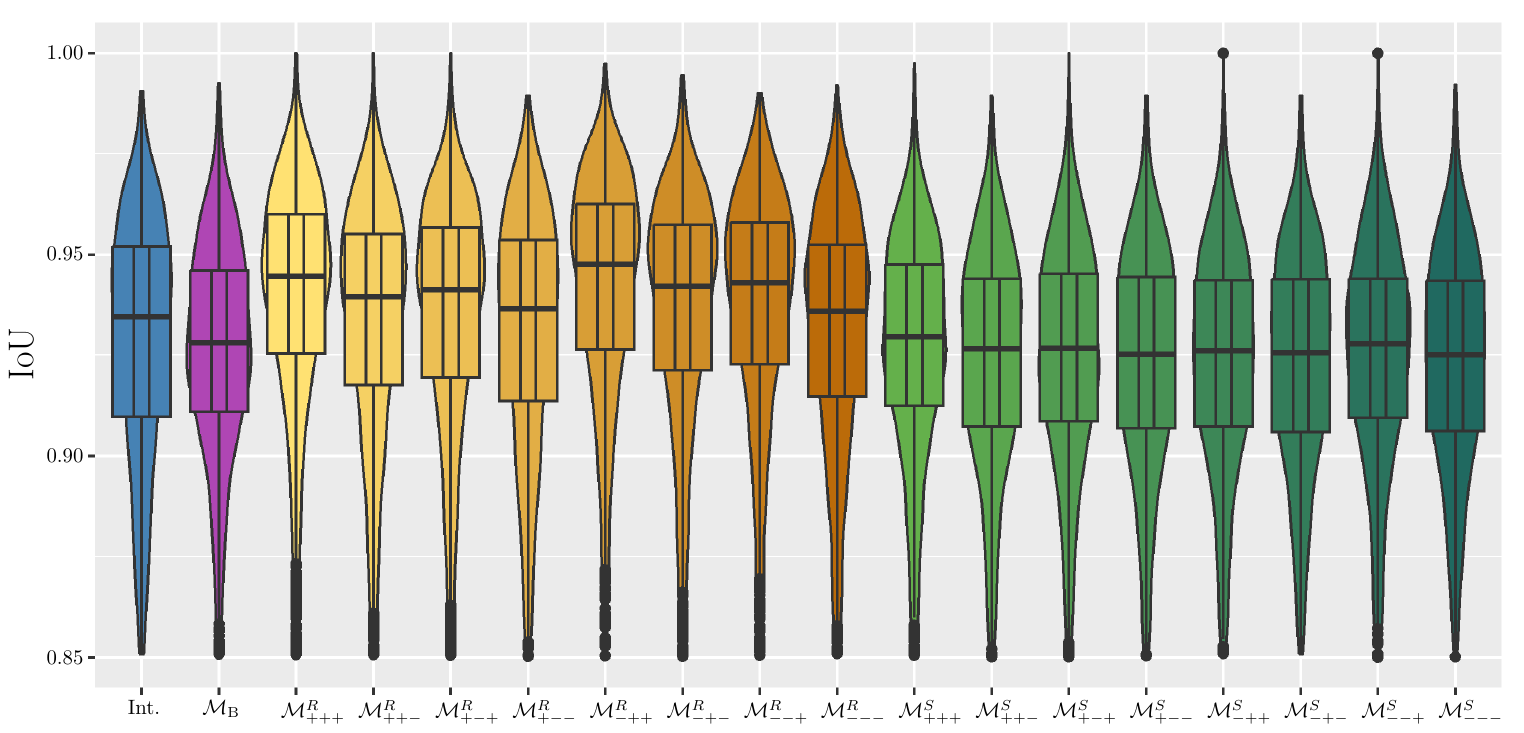}
    \caption{Violin plots illustrating the distribution of the IoU scores for the simulated images obtained with different models.}
    \label{fig:plot_iou_sim}
\end{figure}

\begin{figure}
    \centering
    \includegraphics[width=\textwidth]{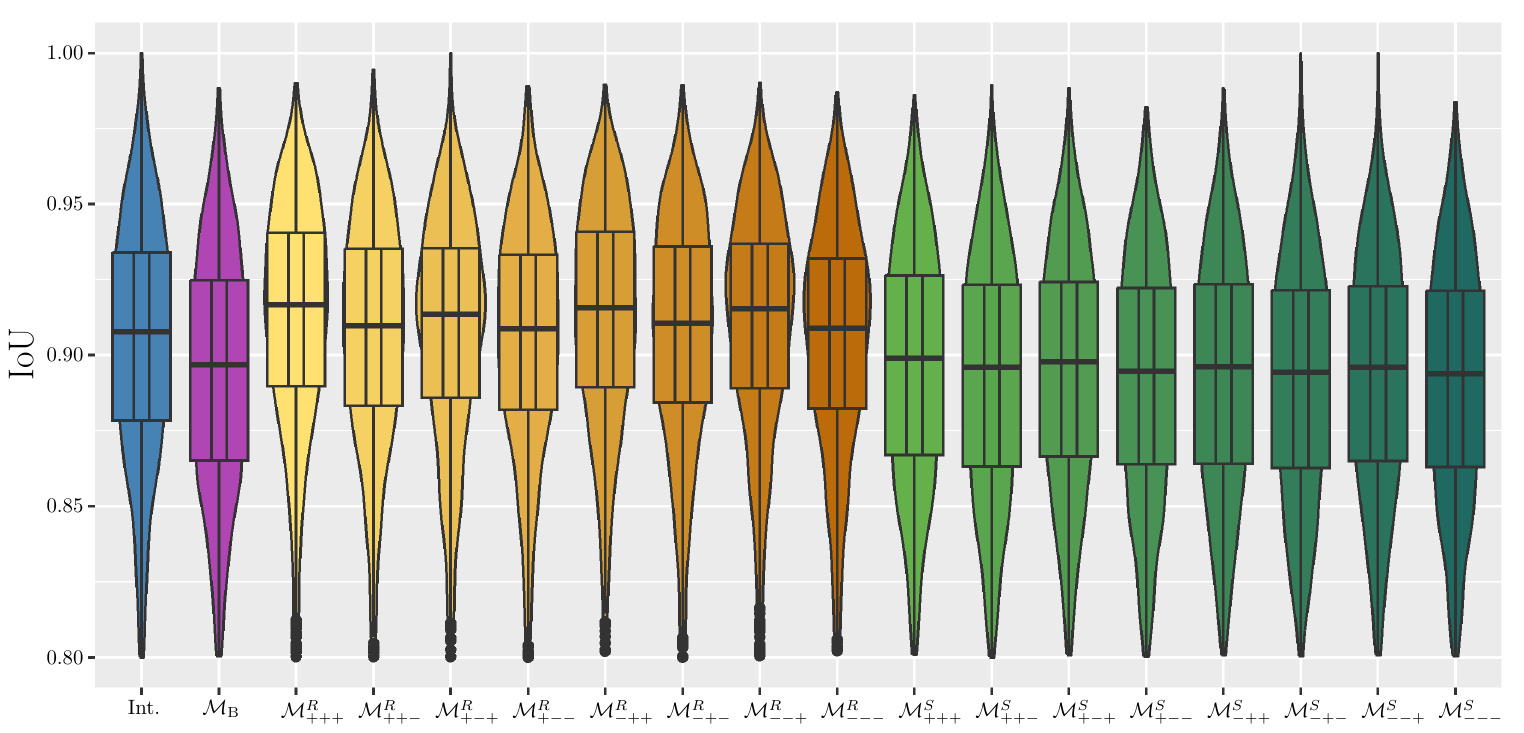}
    \caption{Violin plots illustrating the distribution of the IoU scores for the real-world images obtained with different models.}
    \label{fig:plot_iou_real}
\end{figure}

\definecolor{arsenic}{rgb}{0.23, 0.27, 0.29}
\definecolor{cinnabar}{rgb}{0.89, 0.26, 0.2}
\definecolor{dollarbill}{rgb}{0.52, 0.73, 0.4}

\colorlet{mynegcolor}{cinnabar!40}
\colorlet{myposcolor}{dollarbill!60}
\colorlet{mynscolor}{arsenic!20}

\newcommand{\mytabwidth}{0.95}

\begin{table}[]
\newcommand{\mycellwidth}{9mm}
    \centering

    \caption{Statistical significance of the differences  between the PSNR scores obtained using the investigated models for simulated images. Green color indicates that a variant in the row is significantly better than a variant in the column; red means the opposite. For each row, we present the number of variants that were outperformed (in a statistically-significant way) by the variant in that row.}
    
\resizebox{\mytabwidth\textwidth}{!}{
\begin{tabular}{rwc{\mycellwidth}wc{\mycellwidth}wc{\mycellwidth}wc{\mycellwidth}wc{\mycellwidth}wc{\mycellwidth}wc{\mycellwidth}wc{\mycellwidth}wc{\mycellwidth}wc{\mycellwidth}wc{\mycellwidth}wc{\mycellwidth}wc{\mycellwidth}wc{\mycellwidth}wc{\mycellwidth}wc{\mycellwidth}wc{\mycellwidth}wc{\mycellwidth}r}

\Xhline{2\arrayrulewidth}
\\[-3mm]

&	\cubic & 	\mse & 	\raaa & 	\raab & 	\raba & 	\rabb & 	\rbaa & 	\rbab & 	\rbba & 	\rbbb & 	\saaa & 	\saab & 	\saba & 	\sabb & 	\sbaa & 	\sbab & 	\sbba & 	\sbbb & $\Sigma$ \\ [1mm]	\hline
\rule[4mm]{1mm}{0mm} \cubic & 	\cellcolor{mynegcolor!0} --- & 	\cellcolor{myposcolor!100} *** & 	\cellcolor{myposcolor!100} *** & 	\cellcolor{myposcolor!100} *** & 	\cellcolor{myposcolor!100} *** & 	\cellcolor{myposcolor!100} *** & 	\cellcolor{myposcolor!100} *** & 	\cellcolor{myposcolor!100} *** & 	\cellcolor{myposcolor!100} *** & 	\cellcolor{myposcolor!100} *** & 	\cellcolor{myposcolor!100} *** & 	\cellcolor{myposcolor!100} *** & 	\cellcolor{myposcolor!100} *** & 	\cellcolor{myposcolor!100} *** & 	\cellcolor{myposcolor!100} *** & 	\cellcolor{myposcolor!100} *** & 	\cellcolor{myposcolor!100} *** & 	\cellcolor{myposcolor!100} *** & 	\textbf{17} \\
\rule[4mm]{1mm}{0mm} \mse & 	\cellcolor{mynegcolor!100} *** & 	\cellcolor{mynegcolor!0} --- & 	\cellcolor{myposcolor!100} *** & 	\cellcolor{myposcolor!100} *** & 	\cellcolor{myposcolor!100} *** & 	\cellcolor{myposcolor!100} *** & 	\cellcolor{myposcolor!100} *** & 	\cellcolor{myposcolor!100} *** & 	\cellcolor{myposcolor!100} *** & 	\cellcolor{myposcolor!100} *** & 	\cellcolor{myposcolor!100} *** & 	\cellcolor{myposcolor!100} *** & 	\cellcolor{myposcolor!100} *** & 	\cellcolor{myposcolor!100} *** & 	\cellcolor{myposcolor!100} *** & 	\cellcolor{myposcolor!100} *** & 	\cellcolor{myposcolor!100} *** & 	\cellcolor{myposcolor!100} *** & 	{16} \\
\end{tabular}
    }

    \label{tab:psnr_sim}
\end{table}

\begin{table}[]
\newcommand{\mycellwidth}{9mm}
    \centering

    \caption{Statistical significance of the differences  between the SSIM scores obtained using the investigated models for simulated images. Green color indicates that a variant in the row is significantly better than a variant in the column; red means the opposite. For each row, we present the number of variants that were outperformed (in a statistically-significant way) by the variant in that row.}
    
 \resizebox{\mytabwidth\textwidth}{!}{
     \begin{tabular}{rwc{\mycellwidth}wc{\mycellwidth}wc{\mycellwidth}wc{\mycellwidth}wc{\mycellwidth}wc{\mycellwidth}wc{\mycellwidth}wc{\mycellwidth}wc{\mycellwidth}wc{\mycellwidth}wc{\mycellwidth}wc{\mycellwidth}wc{\mycellwidth}wc{\mycellwidth}wc{\mycellwidth}wc{\mycellwidth}wc{\mycellwidth}wc{\mycellwidth}r}

\Xhline{2\arrayrulewidth}
\\[-3mm]

&	\cubic & 	\mse & 	\raaa & 	\raab & 	\raba & 	\rabb & 	\rbaa & 	\rbab & 	\rbba & 	\rbbb & 	\saaa & 	\saab & 	\saba & 	\sabb & 	\sbaa & 	\sbab & 	\sbba & 	\sbbb & $\Sigma$ \\ [1mm]	\hline
\rule[4mm]{1mm}{0mm} \cubic & 	\cellcolor{mynegcolor!0} --- & 	\cellcolor{myposcolor!100} *** & 	\cellcolor{myposcolor!100} *** & 	\cellcolor{myposcolor!100} *** & 	\cellcolor{myposcolor!100} *** & 	\cellcolor{myposcolor!100} *** & 	\cellcolor{myposcolor!100} *** & 	\cellcolor{myposcolor!100} *** & 	\cellcolor{myposcolor!100} *** & 	\cellcolor{myposcolor!100} *** & 	\cellcolor{myposcolor!100} *** & 	\cellcolor{myposcolor!100} *** & 	\cellcolor{myposcolor!100} *** & 	\cellcolor{myposcolor!100} *** & 	\cellcolor{myposcolor!100} *** & 	\cellcolor{myposcolor!100} *** & 	\cellcolor{myposcolor!100} *** & 	\cellcolor{myposcolor!100} *** & 	\textbf{17} \\
\rule[4mm]{1mm}{0mm} \mse & 	\cellcolor{mynegcolor!100} *** & 	\cellcolor{mynegcolor!0} --- & 	\cellcolor{myposcolor!100} *** & 	\cellcolor{myposcolor!100} *** & 	\cellcolor{myposcolor!100} *** & 	\cellcolor{myposcolor!100} *** & 	\cellcolor{myposcolor!100} *** & 	\cellcolor{myposcolor!100} *** & 	\cellcolor{myposcolor!100} *** & 	\cellcolor{myposcolor!100} *** & 	\cellcolor{myposcolor!100} *** & 	\cellcolor{myposcolor!100} *** & 	\cellcolor{myposcolor!100} *** & 	\cellcolor{myposcolor!100} *** & 	\cellcolor{myposcolor!100} *** & 	\cellcolor{myposcolor!100} *** & 	\cellcolor{myposcolor!100} *** & 	\cellcolor{myposcolor!100} *** & 	{16} \\
\end{tabular}
    }

    \label{tab:ssim_sim}
\end{table}

\begin{table}[]
\newcommand{\mycellwidth}{9mm}
    \centering

    \caption{Statistical significance of the differences  between the LPIPS scores obtained using the investigated models for simulated images. Green color indicates that a variant in the row is significantly better than a variant in the column; red means the opposite. For each row, we present the number of variants that were outperformed (in a statistically-significant way) by the variant in that row.}
    
 \resizebox{\mytabwidth\textwidth}{!}{
     \begin{tabular}{rwc{\mycellwidth}wc{\mycellwidth}wc{\mycellwidth}wc{\mycellwidth}wc{\mycellwidth}wc{\mycellwidth}wc{\mycellwidth}wc{\mycellwidth}wc{\mycellwidth}wc{\mycellwidth}wc{\mycellwidth}wc{\mycellwidth}wc{\mycellwidth}wc{\mycellwidth}wc{\mycellwidth}wc{\mycellwidth}wc{\mycellwidth}wc{\mycellwidth}r}

\Xhline{2\arrayrulewidth}
\\[-3mm]

&	\cubic & 	\mse & 	\raaa & 	\raab & 	\raba & 	\rabb & 	\rbaa & 	\rbab & 	\rbba & 	\rbbb & 	\saaa & 	\saab & 	\saba & 	\sabb & 	\sbaa & 	\sbab & 	\sbba & 	\sbbb & $\Sigma$ \\ [1mm]	\hline
\rule[4mm]{1mm}{0mm} \cubic & 	\cellcolor{myposcolor!0} --- & 	\cellcolor{myposcolor!100} *** & 	\cellcolor{myposcolor!100} *** & 	\cellcolor{myposcolor!100} *** & 	\cellcolor{myposcolor!60} * & 	\cellcolor{myposcolor!100} *** & 	\cellcolor{myposcolor!100} *** & 	\cellcolor{myposcolor!100} *** & 	\cellcolor{myposcolor!100} *** & 	\cellcolor{myposcolor!100} *** & 	\cellcolor{myposcolor!100} *** & 	\cellcolor{myposcolor!100} *** & 	\cellcolor{myposcolor!100} *** & 	\cellcolor{myposcolor!100} *** & 	\cellcolor{myposcolor!100} *** & 	\cellcolor{myposcolor!100} *** & 	\cellcolor{myposcolor!100} *** & 	\cellcolor{myposcolor!100} *** & 	\textbf{17} \\
\rule[4mm]{1mm}{0mm} \mse & 	\cellcolor{mynegcolor!100} *** & 	\cellcolor{myposcolor!0} --- & 	\cellcolor{myposcolor!100} *** & 	\cellcolor{mynscolor} NS & 	\cellcolor{mynegcolor!100} *** & 	\cellcolor{myposcolor!100} *** & 	\cellcolor{myposcolor!100} *** & 	\cellcolor{myposcolor!100} *** & 	\cellcolor{mynscolor} NS & 	\cellcolor{myposcolor!100} *** & 	\cellcolor{mynscolor} NS & 	\cellcolor{mynscolor} NS & 	\cellcolor{myposcolor!60} * & 	\cellcolor{mynscolor} NS & 	\cellcolor{mynscolor} NS & 	\cellcolor{mynscolor} NS & 	\cellcolor{mynegcolor!60} * & 	\cellcolor{mynscolor} NS & 	{6} \\
\end{tabular}
    }

    \label{tab:lpips_sim}
\end{table}

\begin{table}[]
\newcommand{\mycellwidth}{9mm}
    \centering

    \caption{Statistical significance of the differences  between the IoU scores obtained using the investigated models for simulated images. Green color indicates that a variant in the row is significantly better than a variant in the column; red means the opposite. For each row, we present the number of variants that were outperformed (in a statistically-significant way) by the variant in that row.}
    
 \resizebox{\mytabwidth\textwidth}{!}{
     \begin{tabular}{rwc{\mycellwidth}wc{\mycellwidth}wc{\mycellwidth}wc{\mycellwidth}wc{\mycellwidth}wc{\mycellwidth}wc{\mycellwidth}wc{\mycellwidth}wc{\mycellwidth}wc{\mycellwidth}wc{\mycellwidth}wc{\mycellwidth}wc{\mycellwidth}wc{\mycellwidth}wc{\mycellwidth}wc{\mycellwidth}wc{\mycellwidth}wc{\mycellwidth}r}

\Xhline{2\arrayrulewidth}
\\[-3mm]

&	\cubic & 	\mse & 	\raaa & 	\raab & 	\raba & 	\rabb & 	\rbaa & 	\rbab & 	\rbba & 	\rbbb & 	\saaa & 	\saab & 	\saba & 	\sabb & 	\sbaa & 	\sbab & 	\sbba & 	\sbbb & $\Sigma$ \\ [1mm]	\hline
\rule[4mm]{1mm}{0mm} \cubic & 	\cellcolor{mynegcolor!0} --- & 	\cellcolor{mynscolor} NS & 	\cellcolor{myposcolor!80} ** & 	\cellcolor{mynscolor} NS & 	\cellcolor{mynscolor} NS & 	\cellcolor{mynscolor} NS & 	\cellcolor{myposcolor!100} *** & 	\cellcolor{mynscolor} NS & 	\cellcolor{myposcolor!60} * & 	\cellcolor{myposcolor!60} * & 	\cellcolor{mynscolor} NS & 	\cellcolor{mynscolor} NS & 	\cellcolor{mynscolor} NS & 	\cellcolor{mynegcolor!60} * & 	\cellcolor{mynscolor} NS & 	\cellcolor{mynegcolor!60} * & 	\cellcolor{mynscolor} NS & 	\cellcolor{mynegcolor!60} * & 	{4} \\
\rule[4mm]{1mm}{0mm} \mse & 	\cellcolor{mynscolor} NS & 	\cellcolor{mynegcolor!0} --- & 	\cellcolor{myposcolor!100} *** & 	\cellcolor{mynscolor} NS & 	\cellcolor{myposcolor!60} * & 	\cellcolor{mynscolor} NS & 	\cellcolor{myposcolor!100} *** & 	\cellcolor{myposcolor!60} * & 	\cellcolor{myposcolor!80} ** & 	\cellcolor{mynscolor} NS & 	\cellcolor{mynscolor} NS & 	\cellcolor{mynscolor} NS & 	\cellcolor{mynscolor} NS & 	\cellcolor{mynscolor} NS & 	\cellcolor{mynscolor} NS & 	\cellcolor{mynscolor} NS & 	\cellcolor{mynscolor} NS & 	\cellcolor{mynscolor} NS & 	{5} \\
\end{tabular}
    }

    \label{tab:iou_sim}
\end{table}

\begin{table}[]
\newcommand{\mycellwidth}{9mm}
    \centering

    \caption{Statistical significance of the differences  between the PSNR scores obtained using the investigated models for real-world images. Green color indicates that a variant in the row is significantly better than a variant in the column; red means the opposite. For each row, we present the number of variants that were outperformed (in a statistically-significant way) by the variant in that row.}
    
 \resizebox{\mytabwidth\textwidth}{!}{
     \begin{tabular}{rwc{\mycellwidth}wc{\mycellwidth}wc{\mycellwidth}wc{\mycellwidth}wc{\mycellwidth}wc{\mycellwidth}wc{\mycellwidth}wc{\mycellwidth}wc{\mycellwidth}wc{\mycellwidth}wc{\mycellwidth}wc{\mycellwidth}wc{\mycellwidth}wc{\mycellwidth}wc{\mycellwidth}wc{\mycellwidth}wc{\mycellwidth}wc{\mycellwidth}r}

\Xhline{2\arrayrulewidth}
\\[-3mm]

&	\cubic & 	\mse & 	\raaa & 	\raab & 	\raba & 	\rabb & 	\rbaa & 	\rbab & 	\rbba & 	\rbbb & 	\saaa & 	\saab & 	\saba & 	\sabb & 	\sbaa & 	\sbab & 	\sbba & 	\sbbb & $\Sigma$ \\ [1mm]	\hline
\rule[4mm]{1mm}{0mm} \cubic & 	\cellcolor{mynegcolor!0} --- & 	\cellcolor{myposcolor!100} *** & 	\cellcolor{myposcolor!100} *** & 	\cellcolor{myposcolor!100} *** & 	\cellcolor{myposcolor!100} *** & 	\cellcolor{myposcolor!100} *** & 	\cellcolor{myposcolor!100} *** & 	\cellcolor{myposcolor!100} *** & 	\cellcolor{myposcolor!100} *** & 	\cellcolor{myposcolor!100} *** & 	\cellcolor{myposcolor!100} *** & 	\cellcolor{myposcolor!100} *** & 	\cellcolor{myposcolor!100} *** & 	\cellcolor{myposcolor!100} *** & 	\cellcolor{myposcolor!100} *** & 	\cellcolor{myposcolor!100} *** & 	\cellcolor{myposcolor!100} *** & 	\cellcolor{myposcolor!100} *** & 	\textbf{17} \\
\rule[4mm]{1mm}{0mm} \mse & 	\cellcolor{mynegcolor!100} *** & 	\cellcolor{mynegcolor!0} --- & 	\cellcolor{myposcolor!100} *** & 	\cellcolor{mynegcolor!100} *** & 	\cellcolor{mynegcolor!100} *** & 	\cellcolor{myposcolor!100} *** & 	\cellcolor{myposcolor!100} *** & 	\cellcolor{mynscolor} NS & 	\cellcolor{mynegcolor!100} *** & 	\cellcolor{myposcolor!100} *** & 	\cellcolor{myposcolor!100} *** & 	\cellcolor{myposcolor!100} *** & 	\cellcolor{myposcolor!100} *** & 	\cellcolor{myposcolor!100} *** & 	\cellcolor{myposcolor!100} *** & 	\cellcolor{myposcolor!100} *** & 	\cellcolor{mynscolor} NS & 	\cellcolor{myposcolor!100} *** & 	{11} \\
\end{tabular}
    }

    \label{tab:psnr_real}
\end{table}

\begin{table}[]
\newcommand{\mycellwidth}{9mm}
    \centering

    \caption{Statistical significance of the differences  between the SSIM scores obtained using the investigated models for real-world images. Green color indicates that a variant in the row is significantly better than a variant in the column; red means the opposite. For each row, we present the number of variants that were outperformed (in a statistically-significant way) by the variant in that row.}
    
 \resizebox{\mytabwidth\textwidth}{!}{
     \begin{tabular}{rwc{\mycellwidth}wc{\mycellwidth}wc{\mycellwidth}wc{\mycellwidth}wc{\mycellwidth}wc{\mycellwidth}wc{\mycellwidth}wc{\mycellwidth}wc{\mycellwidth}wc{\mycellwidth}wc{\mycellwidth}wc{\mycellwidth}wc{\mycellwidth}wc{\mycellwidth}wc{\mycellwidth}wc{\mycellwidth}wc{\mycellwidth}wc{\mycellwidth}r}

\Xhline{2\arrayrulewidth}
\\[-3mm]

&	\cubic & 	\mse & 	\raaa & 	\raab & 	\raba & 	\rabb & 	\rbaa & 	\rbab & 	\rbba & 	\rbbb & 	\saaa & 	\saab & 	\saba & 	\sabb & 	\sbaa & 	\sbab & 	\sbba & 	\sbbb & $\Sigma$ \\ [1mm]	\hline
\rule[4mm]{1mm}{0mm} \cubic & 	\cellcolor{mynegcolor!0} --- & 	\cellcolor{myposcolor!100} *** & 	\cellcolor{myposcolor!100} *** & 	\cellcolor{myposcolor!100} *** & 	\cellcolor{myposcolor!100} *** & 	\cellcolor{myposcolor!100} *** & 	\cellcolor{myposcolor!100} *** & 	\cellcolor{myposcolor!100} *** & 	\cellcolor{myposcolor!100} *** & 	\cellcolor{myposcolor!100} *** & 	\cellcolor{myposcolor!100} *** & 	\cellcolor{myposcolor!100} *** & 	\cellcolor{myposcolor!100} *** & 	\cellcolor{myposcolor!100} *** & 	\cellcolor{myposcolor!100} *** & 	\cellcolor{myposcolor!100} *** & 	\cellcolor{myposcolor!100} *** & 	\cellcolor{myposcolor!100} *** & 	\textbf{17} \\
\rule[4mm]{1mm}{0mm} \mse & 	\cellcolor{mynegcolor!100} *** & 	\cellcolor{mynegcolor!0} --- & 	\cellcolor{myposcolor!100} *** & 	\cellcolor{mynegcolor!80} ** & 	\cellcolor{mynegcolor!100} *** & 	\cellcolor{myposcolor!100} *** & 	\cellcolor{myposcolor!100} *** & 	\cellcolor{myposcolor!100} *** & 	\cellcolor{mynscolor} NS & 	\cellcolor{myposcolor!100} *** & 	\cellcolor{myposcolor!100} *** & 	\cellcolor{myposcolor!100} *** & 	\cellcolor{myposcolor!100} *** & 	\cellcolor{myposcolor!100} *** & 	\cellcolor{myposcolor!100} *** & 	\cellcolor{myposcolor!100} *** & 	\cellcolor{myposcolor!100} *** & 	\cellcolor{myposcolor!100} *** & 	{13} \\
\end{tabular}
    }

    \label{tab:ssim_real}
\end{table}

\begin{table}[]
\newcommand{\mycellwidth}{9mm}
    \centering

    \caption{Statistical significance of the differences  between the LPIPS scores obtained using the investigated models for real-world images. Green color indicates that a variant in the row is significantly better than a variant in the column; red means the opposite. For each row, we present the number of variants that were outperformed (in a statistically-significant way) by the variant in that row.}
    
 \resizebox{\mytabwidth\textwidth}{!}{
     \begin{tabular}{rwc{\mycellwidth}wc{\mycellwidth}wc{\mycellwidth}wc{\mycellwidth}wc{\mycellwidth}wc{\mycellwidth}wc{\mycellwidth}wc{\mycellwidth}wc{\mycellwidth}wc{\mycellwidth}wc{\mycellwidth}wc{\mycellwidth}wc{\mycellwidth}wc{\mycellwidth}wc{\mycellwidth}wc{\mycellwidth}wc{\mycellwidth}wc{\mycellwidth}r}

\Xhline{2\arrayrulewidth}
\\[-3mm]

&	\cubic & 	\mse & 	\raaa & 	\raab & 	\raba & 	\rabb & 	\rbaa & 	\rbab & 	\rbba & 	\rbbb & 	\saaa & 	\saab & 	\saba & 	\sabb & 	\sbaa & 	\sbab & 	\sbba & 	\sbbb & $\Sigma$ \\ [1mm]	\hline
\rule[4mm]{1mm}{0mm} \cubic & 	\cellcolor{myposcolor!0} --- & 	\cellcolor{myposcolor!100} *** & 	\cellcolor{myposcolor!100} *** & 	\cellcolor{myposcolor!100} *** & 	\cellcolor{myposcolor!100} *** & 	\cellcolor{myposcolor!100} *** & 	\cellcolor{myposcolor!100} *** & 	\cellcolor{myposcolor!100} *** & 	\cellcolor{myposcolor!100} *** & 	\cellcolor{myposcolor!100} *** & 	\cellcolor{myposcolor!100} *** & 	\cellcolor{myposcolor!100} *** & 	\cellcolor{myposcolor!100} *** & 	\cellcolor{myposcolor!100} *** & 	\cellcolor{myposcolor!100} *** & 	\cellcolor{myposcolor!100} *** & 	\cellcolor{myposcolor!100} *** & 	\cellcolor{myposcolor!100} *** & 	\textbf{17} \\
\rule[4mm]{1mm}{0mm} \mse & 	\cellcolor{mynegcolor!100} *** & 	\cellcolor{myposcolor!0} --- & 	\cellcolor{myposcolor!100} *** & 	\cellcolor{myposcolor!60} * & 	\cellcolor{mynegcolor!100} *** & 	\cellcolor{myposcolor!100} *** & 	\cellcolor{myposcolor!100} *** & 	\cellcolor{myposcolor!100} *** & 	\cellcolor{mynscolor} NS & 	\cellcolor{myposcolor!100} *** & 	\cellcolor{myposcolor!100} *** & 	\cellcolor{myposcolor!100} *** & 	\cellcolor{myposcolor!100} *** & 	\cellcolor{myposcolor!100} *** & 	\cellcolor{myposcolor!100} *** & 	\cellcolor{myposcolor!100} *** & 	\cellcolor{myposcolor!80} ** & 	\cellcolor{myposcolor!100} *** & 	{14} \\
\end{tabular}
    }

    \label{tab:lpips_real}
\end{table}

\begin{table}[]
\newcommand{\mycellwidth}{9mm}
    \centering

    \caption{Statistical significance of the differences  between the IoU scores obtained using the investigated models for real-world images. Green color indicates that a variant in the row is significantly better than a variant in the column; red means the opposite. For each row, we present the number of variants that were outperformed (in a statistically-significant way) by the variant in that row.}
    
 \resizebox{\mytabwidth\textwidth}{!}{
     \begin{tabular}{rwc{\mycellwidth}wc{\mycellwidth}wc{\mycellwidth}wc{\mycellwidth}wc{\mycellwidth}wc{\mycellwidth}wc{\mycellwidth}wc{\mycellwidth}wc{\mycellwidth}wc{\mycellwidth}wc{\mycellwidth}wc{\mycellwidth}wc{\mycellwidth}wc{\mycellwidth}wc{\mycellwidth}wc{\mycellwidth}wc{\mycellwidth}wc{\mycellwidth}r}

\Xhline{2\arrayrulewidth}
\\[-3mm]

&	\cubic & 	\mse & 	\raaa & 	\raab & 	\raba & 	\rabb & 	\rbaa & 	\rbab & 	\rbba & 	\rbbb & 	\saaa & 	\saab & 	\saba & 	\sabb & 	\sbaa & 	\sbab & 	\sbba & 	\sbbb & $\Sigma$ \\ [1mm]	\hline
\rule[4mm]{1mm}{0mm} \cubic & 	\cellcolor{mynegcolor!0} --- & 	\cellcolor{myposcolor!100} *** & 	\cellcolor{mynegcolor!100} *** & 	\cellcolor{mynscolor} NS & 	\cellcolor{mynscolor} NS & 	\cellcolor{myposcolor!100} *** & 	\cellcolor{mynegcolor!100} *** & 	\cellcolor{mynscolor} NS & 	\cellcolor{mynegcolor!80} ** & 	\cellcolor{myposcolor!100} *** & 	\cellcolor{myposcolor!100} *** & 	\cellcolor{myposcolor!100} *** & 	\cellcolor{myposcolor!100} *** & 	\cellcolor{myposcolor!100} *** & 	\cellcolor{myposcolor!100} *** & 	\cellcolor{myposcolor!100} *** & 	\cellcolor{myposcolor!100} *** & 	\cellcolor{myposcolor!100} *** & 	{11} \\
\rule[4mm]{1mm}{0mm} \mse & 	\cellcolor{mynegcolor!100} *** & 	\cellcolor{mynegcolor!0} --- & 	\cellcolor{mynegcolor!100} *** & 	\cellcolor{mynegcolor!100} *** & 	\cellcolor{mynegcolor!100} *** & 	\cellcolor{mynegcolor!80} ** & 	\cellcolor{mynegcolor!100} *** & 	\cellcolor{mynegcolor!100} *** & 	\cellcolor{mynegcolor!100} *** & 	\cellcolor{mynscolor} NS & 	\cellcolor{mynscolor} NS & 	\cellcolor{mynscolor} NS & 	\cellcolor{mynscolor} NS & 	\cellcolor{mynscolor} NS & 	\cellcolor{mynscolor} NS & 	\cellcolor{myposcolor!60} * & 	\cellcolor{mynscolor} NS & 	\cellcolor{myposcolor!60} * & 	{2} \\
\end{tabular}
    }

    \label{tab:iou_real}
\end{table}

In Figures~\ref{fig:coco}--\ref{fig:bulletin}, we show the qualitative results. It can be appreciated that the models trained in a task-driven way reconstruct more high-frequency features than the baseline \mse\, configuration, however this is achieved at a cost of some artifacts that affect the image fidelity scores. Also, the outcome obtained with bicubic interpolation is much more blurred compared with all the variants of SR models.
\begin{figure}[!h]
\centering
\includegraphics[width=1.0\textwidth]{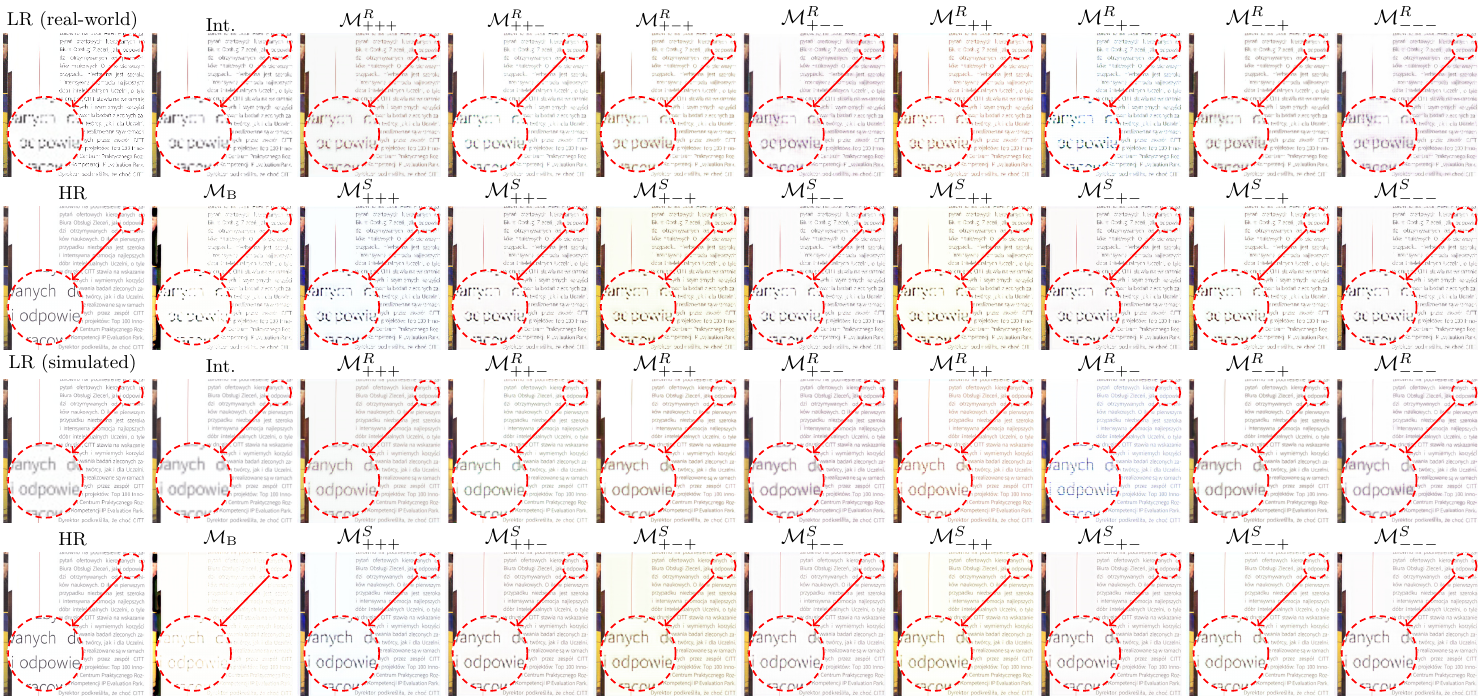}
\caption{Example of reconstructing an image from the University Bulletin dataset performed with SRResNet models trained using different loss functions relying on real-world and simulated training datasets.} \label{fig:bulletin}
\end{figure}
\begin{figure}[!h]
\centering
\includegraphics[width=1.0\textwidth]{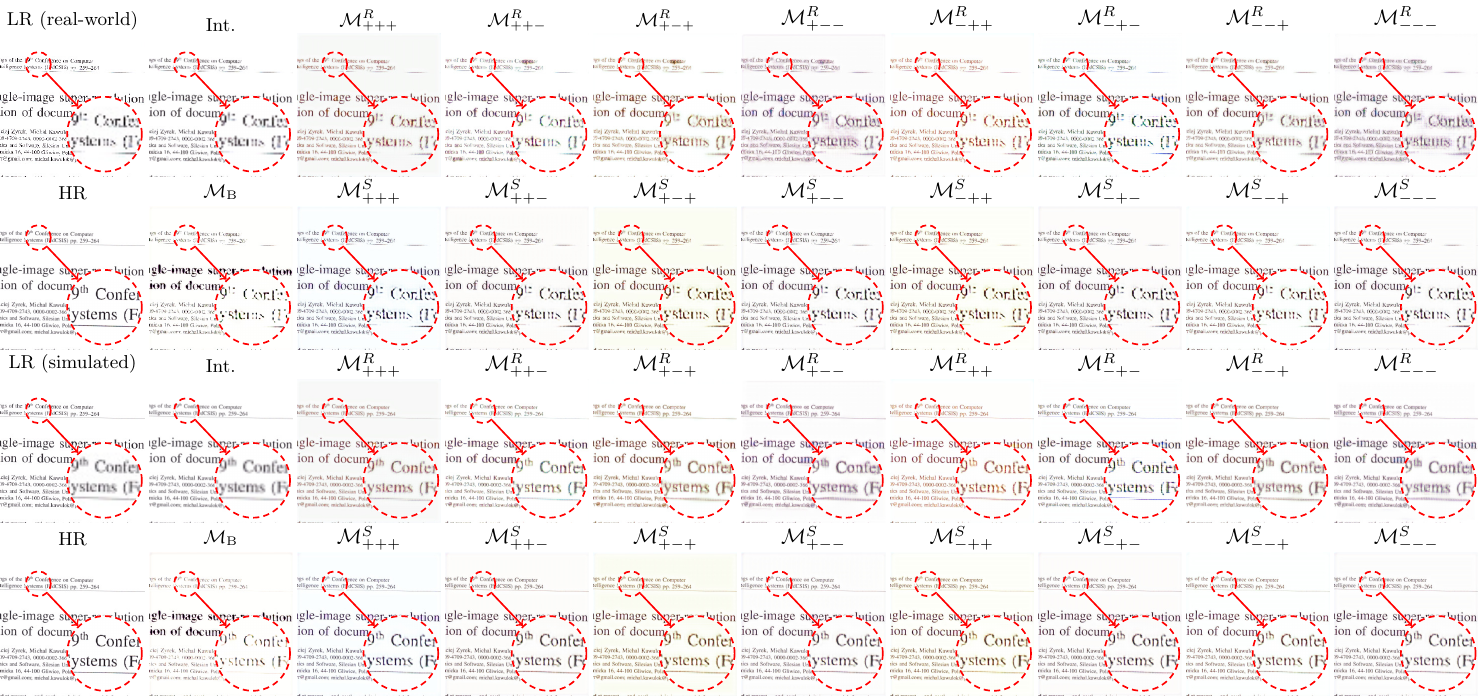}
\caption{Example of reconstructing an image from the Scientific Article dataset performed with SRResNet models trained using different loss functions relying on real-world and simulated training datasets.}
\label{fig:fedcsis}
\end{figure}
\begin{figure}[!h]
\centering
\includegraphics[width=1.0\textwidth]{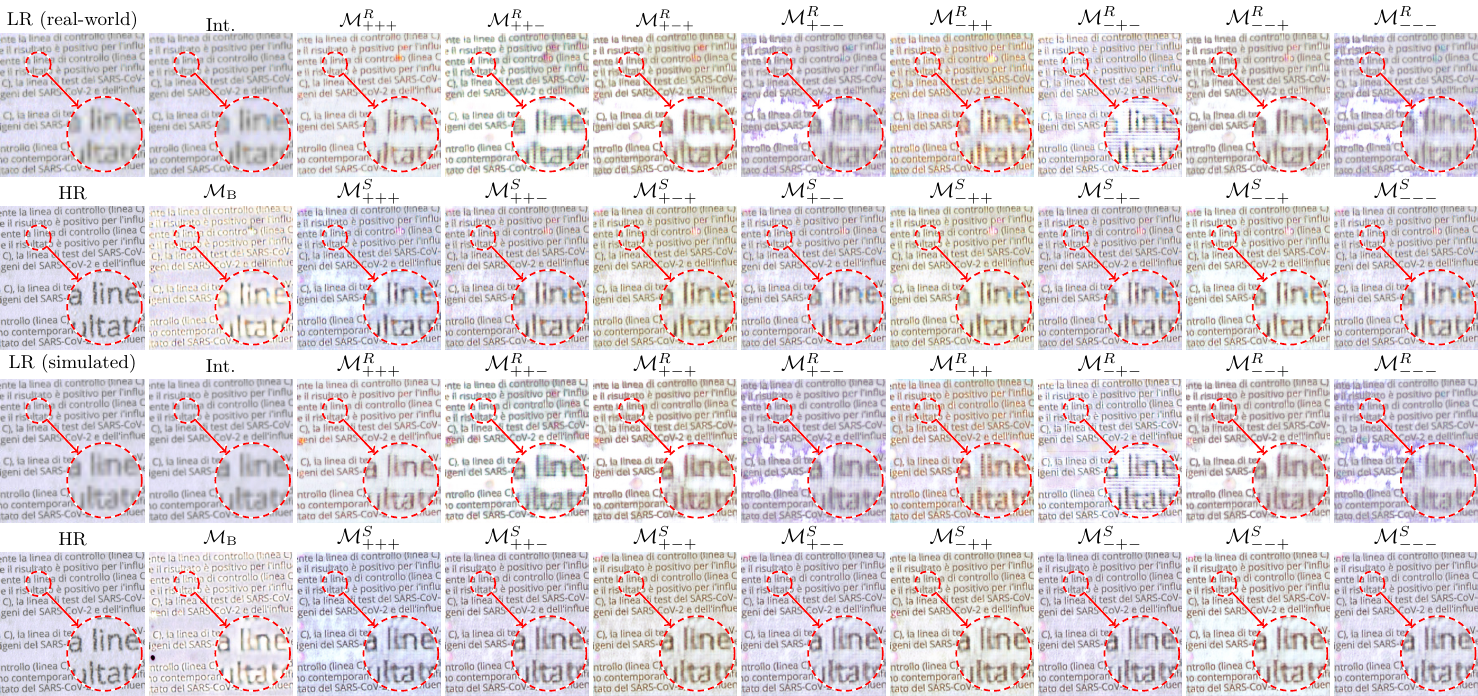}
\caption{Example of reconstructing an image from the Covid Test Leaflet dataset performed with SRResNet models trained using different loss functions relying on real-world and simulated training datasets.} \label{fig:covid}
\end{figure}
\begin{figure}[!h]
\centering
\includegraphics[width=1.0\textwidth]{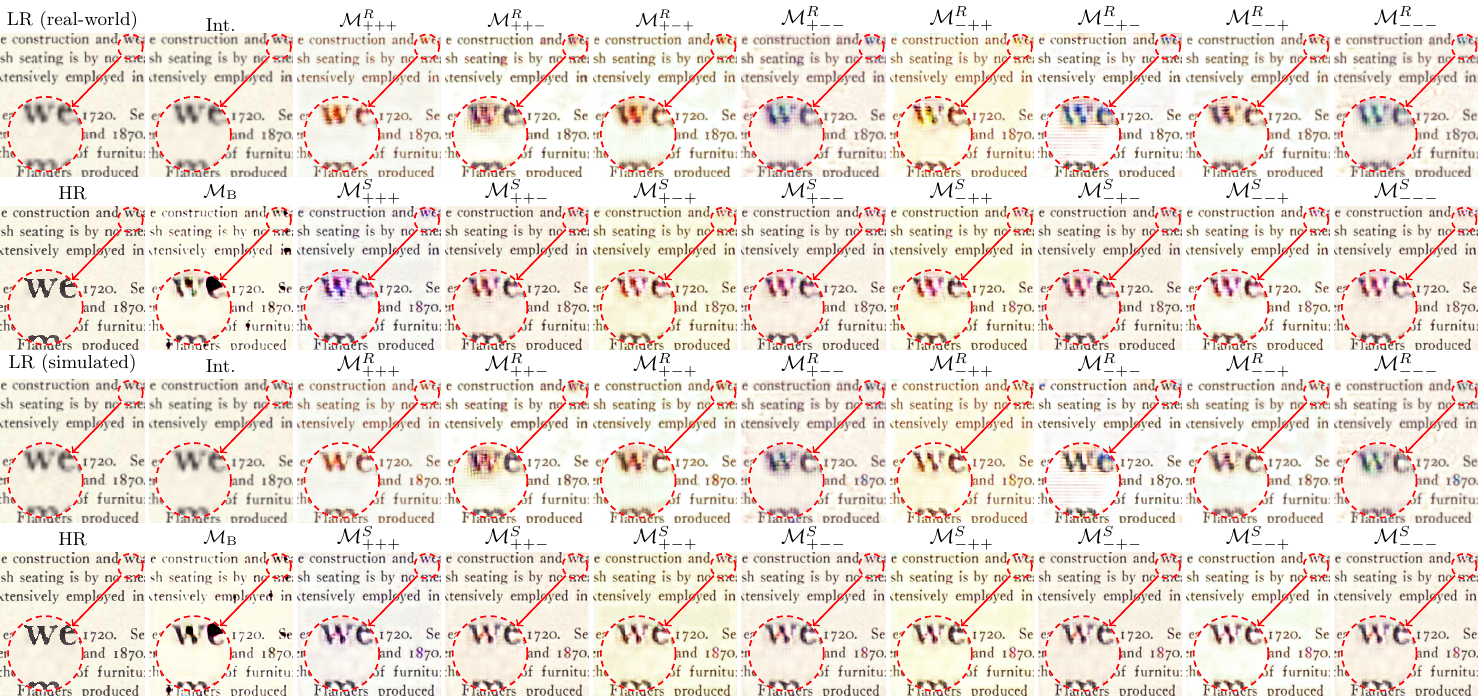}
\caption{Example of reconstructing an image from the Old Books dataset performed with SRResNet models trained using different loss functions relying on real-world and simulated training datasets.}
\label{fig:oldbooks}
\end{figure}
\begin{figure}[!h]
\centering
\includegraphics[width=1.0\textwidth]{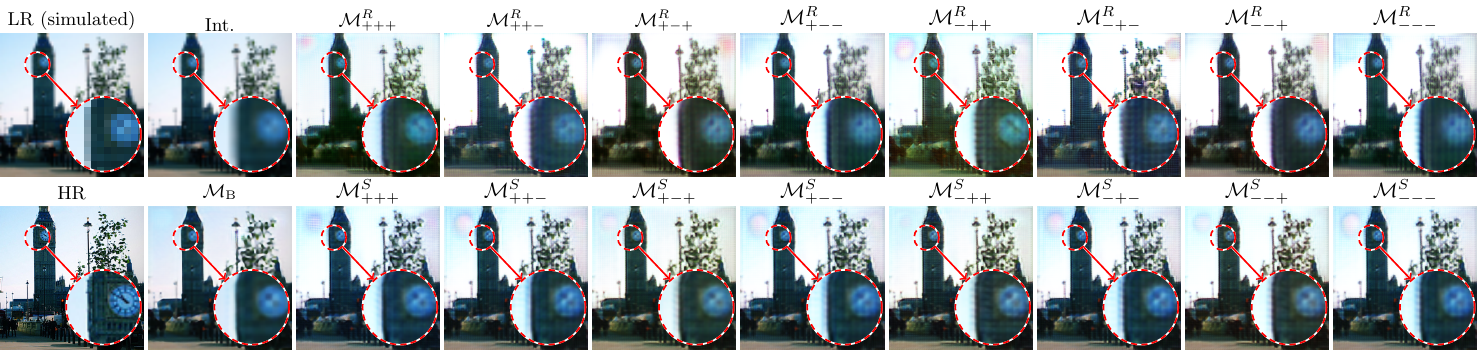}
\caption{Example of reconstructing a natural image from the MS COCO dataset performed with SRResNet models trained using different loss functions relying on real-world and simulated training datasets.} \label{fig:coco}
\end{figure}

\subsection{Discussion} \label{sec:discussion}


Our task-driven multi-task SR approach has clear advantages for OCR-related tasks. By integrating high-level cues, the model learns to preserve or enhance text structures that ordinary SR would blur. For example, faint strokes become bolder (boosting CRNN confidence), and text lines become straighter (raising IoU). This is beneficial in real-world scenarios like receipt processing or historical document analysis, where better OCR accuracy is the goal. Indeed, we find that models optimized for OCR (via CRNN loss) can achieve recognition accuracy on SR images that rivals using the original HR images. However, there are risks of hallucination. SR models can introduce spurious texture or strokes that look plausible but are not present in the real text. We observed some cases (especially under heavy blurring) where the network ``invented'' serifs or glyph fragments, improving IoU metric but potentially altering meaning. Dynamic weighting helps mitigate this: for example, if the CRNN loss overfits to certain text patterns, DWA will reduce its weight as its loss drops. Nevertheless, careful tuning is needed. In practical systems, one might combine SR with a recognition feedback loop to catch inconsistencies. Regarding transferability, our SR model is tailored for OCR on printed text. We tested it briefly on other tasks: for scene text photos (e.g. street signs) it provides moderate gains, since CRNN features are still relevant. But on non-text imagery (natural scenes, faces), its reconstructions degrade, as the model learns biases (e.g. sharp edges, grayscale contrast) beneficial for text but not natural images. This suggests that task-specific SR models should be used judiciously. On the other hand, the multi-task framework itself is general: one could plug in other auxiliary networks (e.g. face detectors, segmentation nets) to train SR for those domains.

\section{Conclusions} \label{sec:concl}

We have presented a multi-task training framework for SRResNet aimed at real-world document scans. By incorporating auxiliary text-detection, recognition, and keypoint alignment networks into the loss, and balancing them via Dynamic Weight Averaging, our model generates super-resolved images that significantly improve OCR-related metrics on challenging scans. Compared to conventional SR trained on synthetic data, our approach better captures the semantics of text content and produces images more amenable to downstream processing. We showed through experiments that the gains are most notable on real scan data, confirming that task-driven SR can bridge the synthetic–real gap. Limitations of our work include the added complexity of multiple networks and the risk of hallucinations. The method also depends on the quality of the auxiliary models (CRNN, CTPN, Key.Net) and their relevance to the data. In future work, we plan to explore end-to-end joint training of the recognition network with SR, larger and more diverse real-image datasets, and online DWA strategies. Extending the framework to other document domains (handwritten text, multi-lingual OCR) and to adversarial settings (improving security) are promising directions.

\section*{Acknowledgements}

This research was funded by the National Science Centre, Poland, under Research Grant no. 2022/47/B/ST6/03009.


\begin{thebibliography}{10}
\providecommand{\url}[1]{\texttt{#1}}
\providecommand{\urlprefix}{URL }
\providecommand{\doi}[1]{https://doi.org/#1}

\bibitem{Ayazoglu2021}
Ayazoglu, M.: Extremely lightweight quantization robust real-time single-image super resolution for mobile devices. In: Proc. IEEE/CVF CVPR. pp. 2472--2479 (2021), \url{https://doi.org/10.1109/CVPRW53098.2021.00280}

\bibitem{Barroso-Laguna_2019_ICCV}
Barroso-Laguna, A., Riba, E., Ponsa, D., Mikolajczyk, K.: Key.net: Keypoint detection by handcrafted and learned cnn filters. In: Proceedings of the IEEE/CVF International Conference on Computer Vision (ICCV) (October 2019)

\bibitem{CaiZeng2019RealSR}
Cai, J., Zeng, H., Yong, H., Cao, Z., Zhang, L.: Toward real-world single image super-resolution: A new benchmark and a new model. In: Proc. IEEE ICCV (2019), \url{https://doi.org/10.1109/ICCV.2019.00318}

\bibitem{ChenHe2022}
Chen, H., He, X., Qing, L., Wu, Y., Ren, C., Sheriff, R.E., Zhu, C.: Real-world single image super-resolution: A brief review. Information Fusion  \textbf{79},  124--145 (2022), \url{https://doi.org/10.1016/j.inffus.2021.09.005}

\bibitem{ChenBadrinarayanan2018}
Chen, Z., Badrinarayanan, V., Lee, C.Y., Rabinovich, A.: {GradNorm}: Gradient normalization for adaptive loss balancing in deep multitask networks. In: Proc. International Conference on Machine Learning (ICML). pp. 794--803 (2018)

\bibitem{Barcha2018}
Correia, P.H.B., Rivera, G.A.R.: Evaluation of {OCR} free software applied to old books. Revista dos Trabalhos de Inicia{\c{c}}{\~a}o Cient{\'\i}fica da UNICAMP (26) (2018)

\bibitem{DengDong2009}
Deng, J., Dong, W., Socher, R., Li, L.J., Li, K., Fei-Fei, L.: {ImageNet}: A large-scale hierarchical image database. In: Proc. Conference on Computer Vision and Pattern Recognition (CVPR). pp. 248--255 (2009)

\bibitem{Dong2014}
Dong, C., Loy, C.C., He, K., Tang, X.: Learning a deep convolutional network for image super-resolution. In: Proc. IEEE/CVF ECCV. pp. 184--199. Springer (2014), \url{https://doi.org/10.1007/978-3-319-10593-2_13}

\bibitem{Dong2015}
Dong, C., Zhu, X., Deng, Y., Loy, C.C., Qiao, Y.: Boosting optical character recognition: A super-resolution approach. arXiv preprint arXiv:1506.02211  (2015), \url{https://doi.org/10.48550/arXiv.1506.02211}

\bibitem{Frizza2022}
Frizza, T., Dansereau, D.G., Seresht, N.M., Bewley, M.: Semantically accurate super-resolution generative adversarial networks. Computer Vision and Image Understanding p. 103464 (2022), \url{https://doi.org/10.1016/j.cviu.2022.103464}

\bibitem{Gomez2017}
Gomez, R., Shi, B., Gomez, L., Numann, L., Veit, A., Matas, J., Belongie, S., Karatzas, D.: {ICDAR2017} robust reading challenge on {COCO}-text. In: 2017 14th IAPR International Conference on Document Analysis and Recognition (ICDAR). vol.~1, pp. 1435--1443. IEEE (2017), \url{https://doi.org/10.1109/ICDAR.2017.234}

\bibitem{GuoDai2023}
Guo, H., Dai, T., Zhu, M., Meng, G., Chen, B., Wang, Z., Xia, S.T.: One-stage low-resolution text recognition with high-resolution knowledge transfer. In: Proc. 31st ACM International Conference on Multimedia. pp. 2189--2198 (2023)

\bibitem{GuoHaque2018}
Guo, M., Haque, A., Huang, D.A., Yeung, S., Fei-Fei, L.: Dynamic task prioritization for multitask learning. In: Proc. European Conference on Computer Vision (ECCV). pp. 270--287 (2018)

\bibitem{Haris2021}
Haris, M., Shakhnarovich, G., Ukita, N.: Task-driven super resolution: Object detection in low-resolution images. In: Proc. ICONIP. pp. 387--395. Springer (2021), \url{https://doi.org/10.1007/978-3-030-92307-5_45}

\bibitem{Honda2022Springer}
Honda, K., Fujita, H., Kurematsu, M.: Improvement of text image super-resolution benefiting multi-task learning. In: Proc. International Conference on Industrial, Engineering and Other Applications of Applied Intelligent Systems. pp. 275--286. Springer (2022)

\bibitem{Kawulok2024JSTARS}
Kawulok, M., Kowaleczko, P., Ziaja, M., Nalepa, J., Kostrzewa, D., Latini, D., De~Santis, D., Salvucci, G., Petracca, I., La~Pegna, V., Bartalis, Z., Del~Frate, F.: Hyperspectral image super-resolution: task-based evaluation. IEEE Journal of Selected Topics in Applied Earth Observations and Remote Sensing  \textbf{17},  18949--18966 (2024)

\bibitem{Kendall2018}
Kendall, A., Gal, Y., Cipolla, R.: Multi-task learning using uncertainty to weigh losses for scene geometry and semantics. In: Proc. Conference on Computer Vision and Pattern Recognition (CVPR). pp. 7482--7491 (2018)

\bibitem{Kim2016}
Kim, J., Kwon~Lee, J., Mu~Lee, K.: Accurate image super-resolution using very deep convolutional networks. In: Proc. IEEE/CVF CVPR. pp. 1646--1654 (2016), \url{https://doi.org/10.1109/CVPR.2016.182}

\bibitem{Ledig2017}
Ledig, C., Theis, L., Husz{\'a}r, F., Caballero, J., Cunningham, A., et~al.: Photo-realistic single image super-resolution using a generative adversarial network. In: Proc. IEEE/CVF CVPR. pp. 4681--4690 (2017), \url{https://doi.org/10.1109/CVPR.2017.19}

\bibitem{LimSon2017}
Lim, B., Son, S., Kim, H., Nah, S., Mu~Lee, K.: Enhanced deep residual networks for single image super-resolution. In: Proc. IEEE/CVF CVPR Workshops. pp. 136--144 (2017), \url{https://doi.org/10.1109/CVPRW.2017.151}

\bibitem{LiuJohns2019}
Liu, S., Johns, E., Davison, A.J.: End-to-end multi-task learning with attention. In: Proc. Conference on Computer Vision and Pattern Recognition (CVPR). pp. 1871--1880 (2019)

\bibitem{LiuWang2023}
Liu, Y., Wang, Y., Shi, H.: A convolutional recurrent neural-network-based machine learning for scene text recognition application. Symmetry  \textbf{15}(4), ~849 (2023)

\bibitem{Liu2019}
Liu, Z., Li, L., Wu, Y., Zhang, C.: Facial expression restoration based on improved graph convolutional networks. arXiv preprint arXiv:1910.10344  (2019)

\bibitem{LuLi2022}
Lu, Z., Li, J., Liu, H., Huang, C., Zhang, L., Zeng, T.: Transformer for single image super-resolution. In: Proc. IEEE/CVF CVPR. pp. 457--466 (2022), \url{https://doi.org/10.1109/CVPRW56347.2022.00061}

\bibitem{Madi2022}
Madi, B., Alaasam, R., El-Sana, J.: Text edges guided network for historical document super resolution. In: International Conference on Frontiers in Handwriting Recognition. pp. 18--33. Springer (2022)

\bibitem{Rad2020}
Rad, M.S., Bozorgtabar, B., Musat, C., Marti, U.V., Basler, M., Ekenel, H.K., Thiran, J.P.: Benefiting from multitask learning to improve single image super-resolution. Neurocomputing  \textbf{398},  304--313 (2020), \url{https://doi.org/10.1016/j.neucom.2019.07.107}

\bibitem{shi2016end}
Shi, B., Bai, X., Yao, C.: An end-to-end trainable neural network for image-based sequence recognition and its application to scene text recognition. IEEE transactions on pattern analysis and machine intelligence  \textbf{39}(11),  2298--2304 (2016)

\bibitem{Simonyan2014}
Simonyan, K., Zisserman, A.: Very deep convolutional networks for large-scale image recognition. arXiv preprint arXiv:1409.1556  (2014)

\bibitem{Vandenhende2021}
Vandenhende, S., Georgoulis, S., Van~Gansbeke, W., Proesmans, M., Dai, D., Van~Gool, L.: Multi-task learning for dense prediction tasks: A survey. IEEE Transactions on Pattern Analysis and Machine Intelligence  \textbf{44}(7),  3614--3633 (2021), \url{https://doi.org/10.1109/TPAMI.2021.3054719}

\bibitem{WanYu2024}
Wan, C., Yu, H., Li, Z., Chen, Y., Zou, Y., Liu, Y., Yin, X., Zuo, K.: Swift parameter-free attention network for efficient super-resolution. In: Proc. IEEE/CVF Conference on Computer Vision and Pattern Recognition. pp. 6246--6256 (2024)

\bibitem{WangXie2019}
Wang, W., Xie, E., Sun, P., Wang, W., Tian, L., Shen, C., Luo, P.: {TextSR}: Content-aware text super-resolution guided by recognition. arXiv preprint arXiv:1909.07113  (2019), \url{https://doi.org/10.48550/arXiv.1909.07113}

\bibitem{XieZhang2023}
Xie, C., Zhang, X., Li, L., Meng, H., Zhang, T., Li, T., Zhao, X.: Large kernel distillation network for efficient single image super-resolution. In: Proceedings of the IEEE/CVF conference on computer vision and pattern recognition. pp. 1283--1292 (2023)

\bibitem{YuShi2024}
Yu, M., Shi, J., Xue, C., Hao, X., Yan, G.: A review of single image super-resolution reconstruction based on deep learning. Multimedia Tools and Applications  \textbf{83}(18),  55921--55962 (2024)

\bibitem{Zhang2018}
Zhang, Y., Tian, Y., Kong, Y., Zhong, B., Fu, Y.: Residual dense network for image super-resolution. In: Proc. IEEE/CVF CVPR. pp. 2472--2481 (2018)

\bibitem{zyrek2024task}
Zyrek, M., Kawulok, M.: Task-driven single-image super-resolution reconstruction of document scans. In: 2024 19th Conference on Computer Science and Intelligence Systems (FedCSIS). pp. 259--264. IEEE (2024)

\end{thebibliography}
\end{document}